\documentclass[10pt,twocolumn,letterpaper]{article}

\usepackage{cvpr}              

\usepackage{graphicx}
\usepackage{amsmath}
\usepackage{amssymb}
\usepackage{float}
\usepackage{booktabs}
\usepackage{bbm}
\usepackage{pifont}

\usepackage[pagebackref,breaklinks,colorlinks]{hyperref}

\usepackage[capitalize]{cleveref}
\crefname{section}{Sec.}{Secs.}
\Crefname{section}{Section}{Sections}
\Crefname{table}{Table}{Tables}
\crefname{table}{Tab.}{Tabs.}

\def\cvprPaperID{*****} 
\def\confName{CVPR}
\def\confYear{2023}

\begin{document}


\def\paperTitle{Neuromorphic Event-based Facial Expression Recognition} 
\def\authorBlock{ 
Lorenzo Berlincioni\thanks{Equal contribution} $^1$~~~
\quad Luca Cultrera\footnotemark[1] $^1$~~~
\quad Chiara Albisani$^1$~~~
\quad Lisa Cresti$^1$~~~
\quad Andrea Leonardo$^1$~~~\\
\quad Sara Picchioni$^1$~~~
\quad Federico Becattini$^2$~~~
\quad Alberto Del Bimbo$^1$~~~

\\
$^1$University of Florence, MICC {\tt\small \{name.surname\}@unifi.it}
\\
$^2$University of Siena {\tt\small \{name.surname\}@unisi.it}

 }
\author{\authorBlock}
\title{Neuromorphic Event-based Facial Expression Recognition}









\maketitle

\begin{abstract}
Recently, event cameras have shown large applicability in several computer vision fields especially concerning tasks that require high temporal resolution. In this work, we investigate the usage of such kind of data for emotion recognition by presenting NEFER, a dataset for Neuromorphic Event-based Facial Expression Recognition. NEFER is composed of paired RGB and event videos representing human faces labeled with the respective emotions and also annotated with face bounding boxes and facial landmarks. We detail the data acquisition process as well as providing a baseline method for RGB and event data. The collected data captures subtle micro-expressions, which are hard to spot with RGB data, yet emerge in the event domain. We report a double recognition accuracy for the event-based approach, proving the effectiveness of a neuromorphic approach for analyzing fast and hardly detectable expressions and the emotions they conceal.
\end{abstract}

\section{Introduction}
Facial expression recognition is important for a large variety of applications \cite{li2020deep,guerdelli2022macro,becattini2021plm}. Different kinds of sensors have been used to analyze faces such as depth cameras \cite{corneanu2016survey} or sensors with high framerate such as high-speed structured light sensors \cite{ye2021facial} and extremely fast RGB cameras \cite{polikovsky2009facial}. In particular, the necessity for elevated framerates stems from the fact that emotions are often conveyed by micro-expressions, which can manifest in short timespans up to 1/25 of a second \cite{ekman2009telling}.
Recently, an exploratory approach has studied the capability of neuromorphic sensors, i.e. event cameras, to capture facial expressions \cite{becattini2022understanding}. It suggested better recognition rates for event-based approaches compared to RGB.

\begin{figure}[h]
\begin{center}
  \includegraphics[width=.95\linewidth]{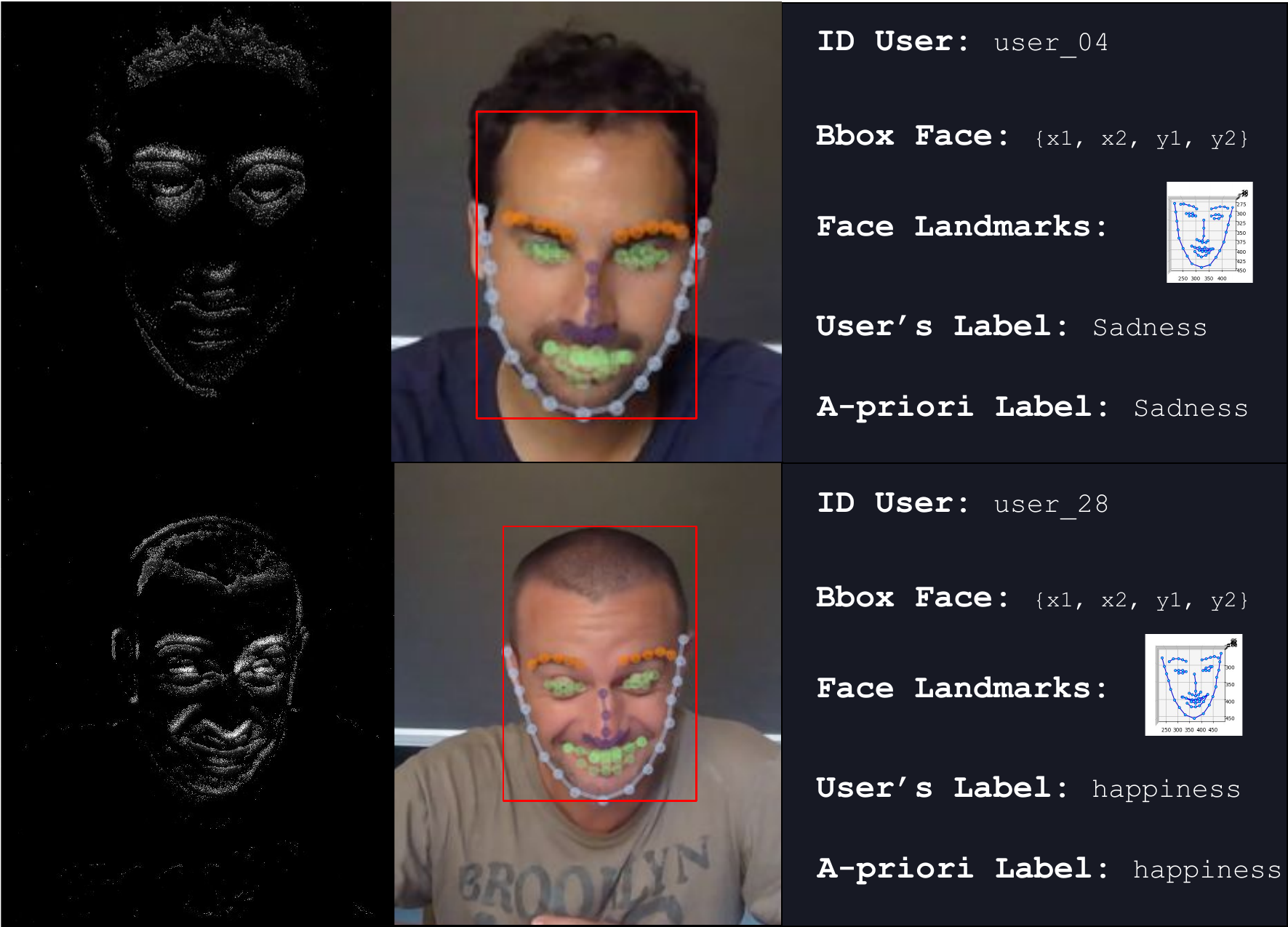}
\end{center}
  \caption{NEFER is a dataset for Neuromorphic Event-based Facial Expression Recognition. We collect paired Event streams and RGB videos, providing for both modalities face bounding boxes, facial landmarks and emotion labels. Emotion labels are provided in two versions: using an a-priori assignment based on the visual stimulus shown to the user and based on actual user feelings.}
\label{fig:teaser}
\end{figure}

Event cameras are bio-inspired sensors that, instead of generating streams of synchronous frames, produce asynchronous events for single pixels where illumination changes occur. An advantage is the extremely high rate of events, with temporal resolutions that reach the microsecond.
However, due to a lack of data, emotion recognition through event-based videos is still a problem not widely addressed in the literature. 
In order to cope with the aforementioned lack of data, several attempts have been made to generate synthetic event-based datasets \cite{Rebecq18corl, joubert2021EventCS, hu2021v2eFV}.
The authors of \cite{becattini2022understanding} framed the recognition setting as a facial reaction recognition system, aiming at understanding whether an expression is positive or negative when using an interactive recommendation system.
The authors however, collected a dataset using a VGA event camera, thus with limited spatial resolution. We believe that this poses a strong limit for facial-expression analysis applications since micro-expressions can be very localized in space as much as in time.
Reactions are also labeled just as positive, neutral and negative, without providing details about emotions.
A few additional works have addressed similar problems, yet focusing only on face detection and tracking alone, without analyzing expressions or emotions \cite{savran2020face,lenz2020event}.

In our work we present NEFER (Neuromorphic Event-based Facial Expression Recognition)
\footnote{The dataset is available here: \url{https://github.com/miccunifi/NEFER}}
, the first release of an RGB and event dataset for emotion recognition. 
The dataset is fully labelled with bounding boxes from face detection and facial landmark.

As traditional annotation methods are not ideal for event-based data, we chose a hybrid approach. This involved using the ESIM \cite{Rebecq18corl} simulator to obtain aligned RGB images and event streams, which we could then analyze using supervision signals obtained directly from RGB vision methods for face detection and face landmark estimation.
We also provide a simple baseline to underline the difficulty of the task and the capabilities of an event-based model with reference to an RGB counterpart. To the best of our knowledge, we are the first to publicly release an event camera facial expression recognition dataset.
To summarize, the main contributions of this paper are:
\begin{itemize}
    \item We propose a dataset for  emotion recognition  recorded with an high resolution event camera. The  dataset consists of more than 600 RGB and     events-based videos from more than 30 individuals of different genders and ages.
    \item We provide labels for face and landmark detection in both RGB and event data.
    \item For each sample in the dataset we provide two different kind of labels: a-priori labels (pre-defined emotion assignment) and user labels (emotion felt by the user).
    \item We provide a baseline model to foster future research in the field, underlying the potential of event-based analysis compared to standard RGB approaches.

\end{itemize}

\newcommand{\cmark}{\ding{51}}%
\newcommand{\xmark}{\ding{55}}%
\begin{table*}[]
    \centering
    \begin{tabular}{l|c|c|c|c|c|c}
       Dataset & Videos & Users & Resolution & Bounding Boxes & Landmarks & Emotions\\ \hline
       Savran \textit{et al.} \cite{savran2020face} & 108 & \textbf{30} & 304 $\times$ 204 & \xmark & \xmark & \xmark \\
       Lenz \textit{et al.} \cite{lenz2020event} & 48 & 10 & 640 $\times$ 480 & \xmark & \xmark & \xmark \\
       Becattini \textit{et al.} \cite{becattini2022understanding} & 455 & 25 & 640 $\times$ 480 & \xmark & \xmark & \xmark \\
       NEFER & \textbf{609} & 29 & \textbf{1280 $\times$ 720} & \cmark & \cmark & \cmark \\
    \end{tabular}
    \caption{Comparison of event-based face datasets. To the best of our knowledge, NEFER is the first dataset to provide bounding boxes, facial landmarks and emotion labels, as well as an HD resolution. NEFER is also the larger face dataset up to date.}
    \label{tab:dataset_comparison}
\end{table*}

\section{Related Works}
The event camera is a neuromorphic sensor that is based on a novel bio-inspired vision paradigm \cite{delbruckl2016neuro, posch2014retino}. In contrast to traditional vision systems, it does not produce a synchronous sequence of frames, but instead generates an asynchronous stream of events. Each event is characterized by a local change in brightness and can occur at very short time intervals (in the order of microseconds) with very low latency \cite{lichtsteiner2008asynch}. Moreover, unlike traditional vision systems, the event camera does not produce any output if there is no change in brightness, thereby conserving resources. 
To summarize, utilizing a neuromorphic sensor results in reduced motion blur, high temporal resolution, and high dynamic range (up to 140 dB). Additionally, it enables a reduction in bandwidth consumption \cite{finateu2020back, gallego2020event}.

Despite the fact that event cameras have not been on the market for an extended period, and their large-scale use is still somewhat limited, there are examples of their application in fields such as robotics and computer vision that can be found in the literature \cite{gallego2020event,delbruckl2016neuro}. In fact, in these contexts, the benefits offered by event cameras can be fully leveraged.
In \cite{ramesh2019dart} the authors propose an event-based descriptor for event camera data and show its results in some vision problems such as object classification, tracking, detection and feature matching.
Also \cite{mondal2021moving, li2021graph, kim2021n } propose event-based approaches for object detection and recognition.
Event cameras are widely used in literature for tracking \cite{renner2020event, seok2020robbust, zhang2021object}.
In \cite{zhang2022spiking} they propose a trasformer-based architecture to fuse temporal and spatial information encoded in the events for single object tracking.
Neuromorphic sensors are extensively utilized in surveillance \cite{rodriguez2020async,sultana2019iot, litzenberger2006estimation } due to their distinctive characteristics and low power consumption. In fact, one of the most desirable properties of these sensors in surveillance is their ability to transmit information solely when changes occur.
\cite{chen2022neuro} propose a neuromorphic vision-based system for autonomous vehicles. 
Event cameras are also used in a wide range of scenarios in robotics and computer vision such as video super-resolution~\cite{jing2021turning, han2021evintsr}, depth and optical flow prediction \cite{gallego2019focus}, monocular and stereo depth estimation \cite{gehrig2021combining, uddin2022unsupervised}, SLAM \cite{jiao2021comparing} and visual odometry \cite{zuo2022devo, wang2022visual, liu2021spatiotemporal}, and
human pose estimation \cite{scarpellini2021lifting, chen2022efficient}.

Of particular interest for this work, is the fact that neuromorphic sensors have a compelling application scenario in face detection and emotion recognition. The distinct characteristics and properties of event cameras enable them to capture even the subtlest variations and microexpressions in human emotions at remarkably high temporal resolution and with minimal latency. Nevertheless, this aspect has not received widespread attention in the literature. 
In general, facial images possess crucial features that can serve several biometric applications \cite{mehdipour2016comprehensive} and therefore face and landmark detection through deep learning algorithms is a problem widely addressed in the literature \cite{wen2016discriminative,wu2019facial,taskiran2020face}.
Training deep learning models to perform well in face and landmark detection tasks requires a large amount of data: \cite{yi2014learning} proposes a dataset composed by 0.5M images from 10,575 individuals, \cite{guo2016ms} use more than 100k individuals to generate approximately 10M of images, and \cite{kemelmacher2016megaface} includes 1M images from more the 690k individuals. The usage of synthetic face images has also been explored \cite{gecer2018semi}, whereas \cite{wang2020masked} instead proposed a dataset for masked face recognition, as a response to safety mandates during the covid pandemic.
Several other datasets have also been published addressing the study of faces \cite{cao2018vggface2, wang2018devil, huang2008labeled}.

As for the event-based domain, despite all this interest in the topic, not many datasets can be found in the literature. In fact in \cite{ryan2021real}, to make up for the lack of data, they use a synthetic event-based dataset starting from \cite{le2012interactive}.
Several attempts have been made in the literature to generate simulated data for event cameras. In \cite{Rebecq18corl} the authors propose ESIM, an event camera simulator with the ability to accurately and efficiently simulate events, while also offering the flexibility to simulate any camera trajectory within a 3D scene of any nature. \cite{joubert2021EventCS} extends \cite{Rebecq18corl} with the goal to reduce the gap between simulation and real sensors by directly mapping noise distributions from real pixels. \cite{hu2021v2eFV}~instead, proposes a tool for generating synthetic event data and demonstrates its effectiveness in two computer vision object recognition and detection tasks. 
Also \cite{ziegler2022realtime} and \cite{pantho2021EventCS} propose event camera simulators.
On the one hand, \cite{ziegler2022realtime} introduces a multiple event simulator method suitable to be used in real-time robotics applications.
\cite{pantho2021EventCS}, on the other hand, suggest an approach to emulate the behavior of an attention-based camera sensor.
In conclusion, to the best of our knowledge, only three datasets containing facial images captured using a real event camera are present in the literature \cite{savran2020face, lenz2020event, becattini2022understanding}.
In \cite{savran2020face} the problem of face pose alignment is analyzed. The authors provide a dataset consisting of 108 videos of extreme head rotations with varying motion intensity, totaling just over 10 minutes of frames acquired.
In \cite{lenz2020event} on the other hand, the authors collected data with an event camera for eye blink detection. The dataset consists of 48 videos (total duration of about 13 minutes).
The authors of \cite{becattini2022understanding} instead collected a dataset of 455 videos of facial reactions where the recorded users react to garment images. Reactions are classified in three classes: positive, neutral and negative.
However, both \cite{savran2020face}, \cite{lenz2020event} and \cite{becattini2022understanding} use low resolution event cameras with resolution of $304 \times 240$px or $640 \times 480$px. We are of the opinion that this presents a significant constraint for facial expression analysis applications as micro-expressions can be highly localized both spatially and temporally. 

In this paper, we introduce NEFER (Neuromorphic Event-based Facial Expression Recognition), a dataset composed of paired RGB and event data for emotion recognition.
We collected the dataset with high-resolution RGB and event cameras, providing also facial bounding box and landmark annotations in addition to emotion labels following the frequently used Ekman's emotion classification \cite{ekmann1973universal}.
As far as our knowledge extends we are the first to publicly release an event camera-based facial expression recognition dataset. A comparison with existing datasets is presented in Tab.~\ref{tab:dataset_comparison}.

\begin{table*}[t]
\resizebox{\textwidth}{!}{%
\begin{tabular}{c|ccc} 
\hline
 Emotion & & Video Descriptions &\\ 
 \hline \hline
  Disgust & Spyder in a man's mouth &  Crushed Pimple on Cheek & Man Eating a Larva \\ \hline
 
 Contempt & Cops Killing Protestant & Dog Being Abandoned & Dog Being Mugged \\ \hline
 
 Happiness & Dogs playing & Laughing Child & Old Man Dancing with Boys  \\ \hline
 
 Fear & Suddenly Appearing Ghost & Hidden Clown Attacking Camera & Giant Snake Attacking Camera \\ \hline
 
 Anger & Man Attacking Companion & Boy destroying Brother's PC & Professor Assaulted by Student's Parents \\ \hline
 
 Surprise & Baseball Coming Towards Camera & Girl with Unexpected Makeup & Presentation Concluding with a Cat \\ \hline
 
 Sadness & Death of Mufasa in the Lion King & Death of Elly in Up & Boy who has to Undergo an Operation \\ \hline
 \end{tabular}}
 \caption{Videos shown to participants and relative emotion label. Emotions follow the 7 basic emotions defined by Ekman \cite{ekmann1973universal}. \label{tab:videos}}
\end{table*}

\begin{figure}[t]
\begin{center}
  \includegraphics[width=.95\linewidth]{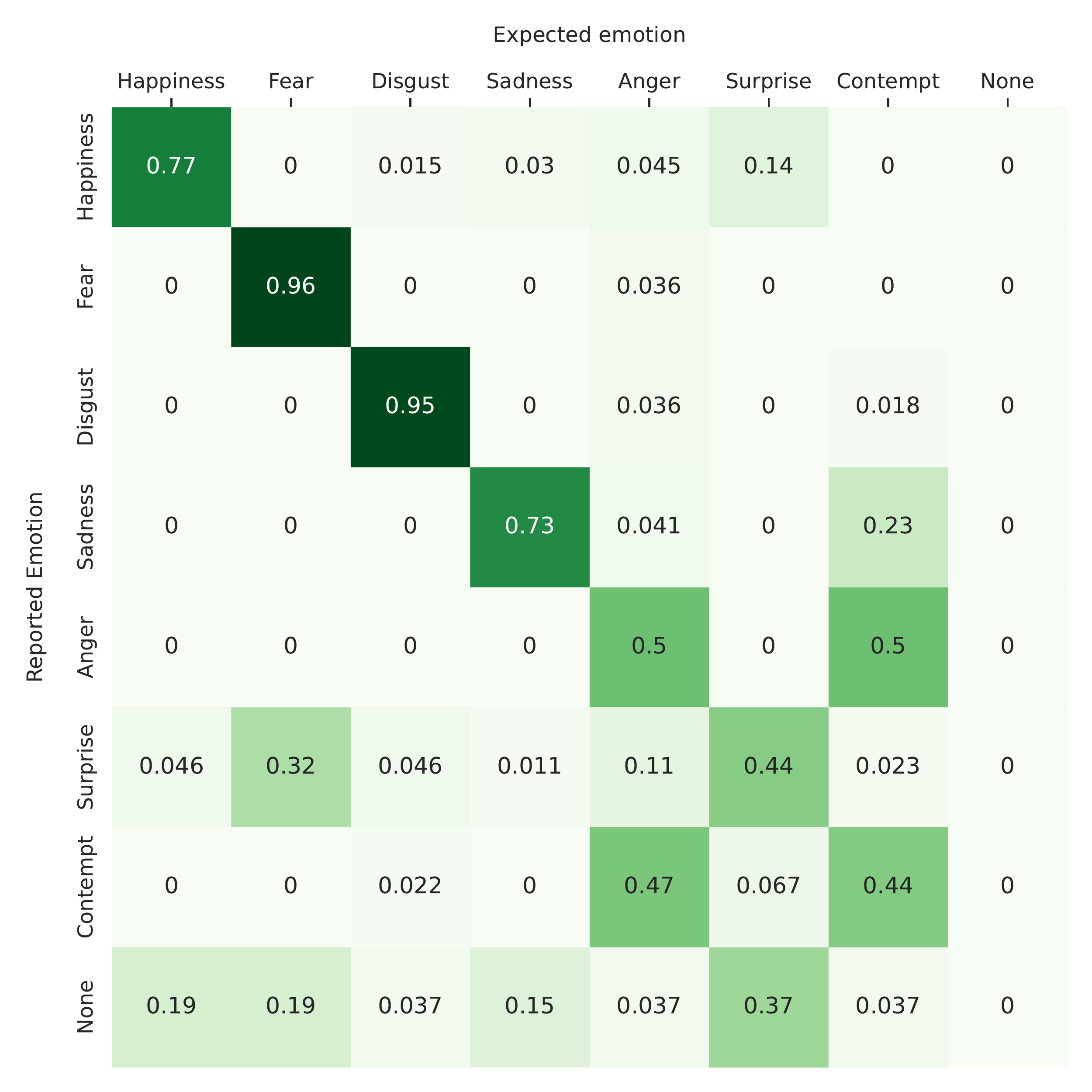}
\end{center}
  \caption{Confusion matrix of the two sets of labels from the NEFER dataset. The expected emotions are a-priori labels assigned based on the content of the visual stimulus shown to the users. The reported emotions instead are emotions declared by the users after observing the videos.}
\label{fig:cf_matrix}
\end{figure}

\section{NEFER: Neuromorphic Event-based Facial Expression Recognition}
The purpose of NEFER is to capture genuine micro-expressions associated to specific emotions with both an event camera and a standard RGB camera.
We considered the 7 primary emotions defined by Ekman \cite{ekmann1973universal}, namely \textit{Disgust}, \textit{Contempt}, \textit{Happiness}, \textit{Fear}, \textit{Anger}, \textit{Surprise} and \textit{Sadness}, since these have been identified as independent from culture, history and personality and are performed in a similar way by everyone.
\subsection{Setting and protocol}
In order to obtain realistic and non-simulated expressions, we asked a set of volunteers to maintain a neutral facial expression while watching a selection of videos. A reward has been offered to the participants to encourage a proper behavior during the test (high-stakes situation).
The volunteers that took part in the creation of NEFER are both males and females of age ranging between 24 and 52 years, for a total of 29 users.

We showed to each user 21 different videos, 3 for each of Ekman's basic emotions. The videos have been selected from online streaming platforms (e.g. YouTube). Each video was trimmed to the same length of 7s to keep the recording sessions as short as possible so not to induce unwanted expressions due to, for instance, boredom.
This choice also simplifies training schemes with deep learning frameworks which process data in mini-batches of the same size. Tab.~\ref{tab:videos} lists the videos that have been shown to the users with the correspondent emotion label. The overall procedure for the data acquisition and video selection was inspired by previously collected dataset from the state of the art \cite{yan2013casme,davison2016samm}.

 
 
 
 
 
 

\begin{figure}[t]
	\begin{center}
		\includegraphics[width=\columnwidth]{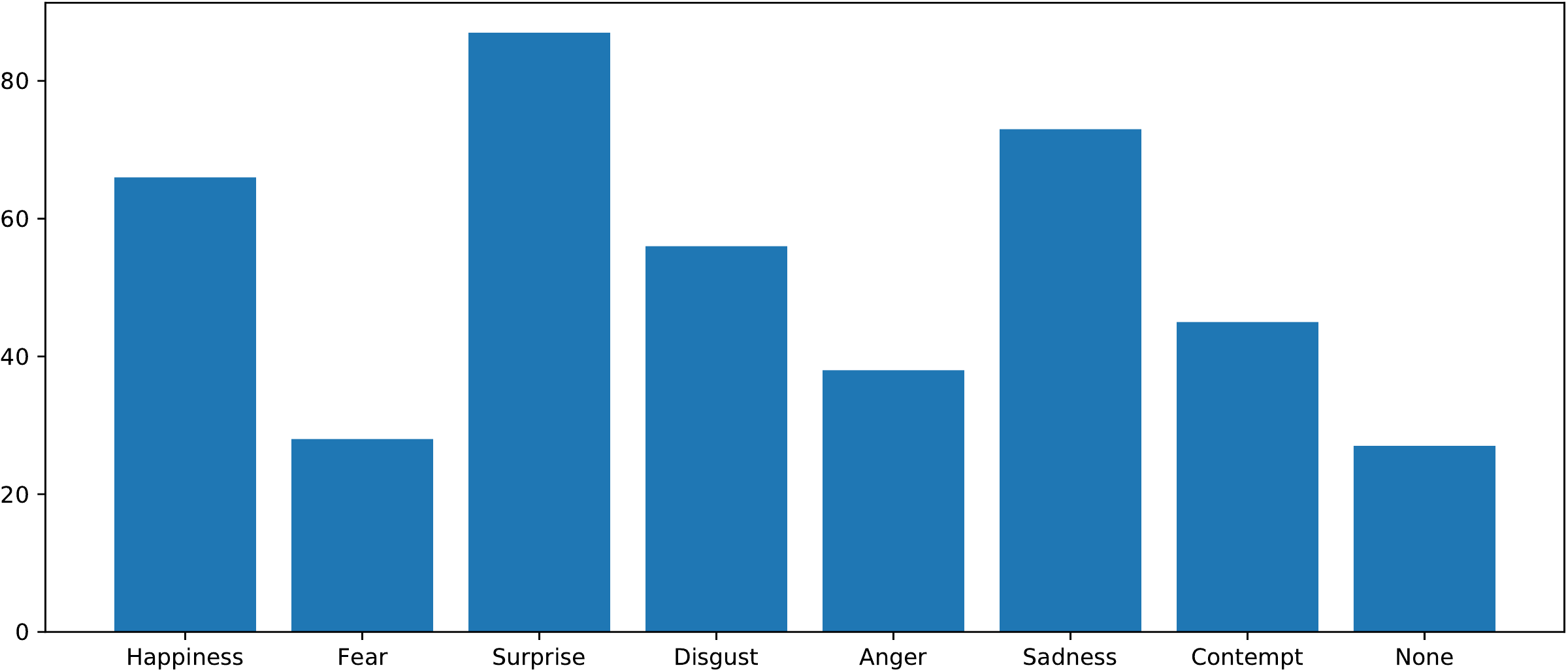}
    \end{center}
	\caption{Class distribution with user labels.}
	\label{fig:distribution_labels}
\end{figure}

\newcommand{\myw}{0.32\columnwidth}
\begin{figure*}
	\begin{center}
		\includegraphics[width=\myw]{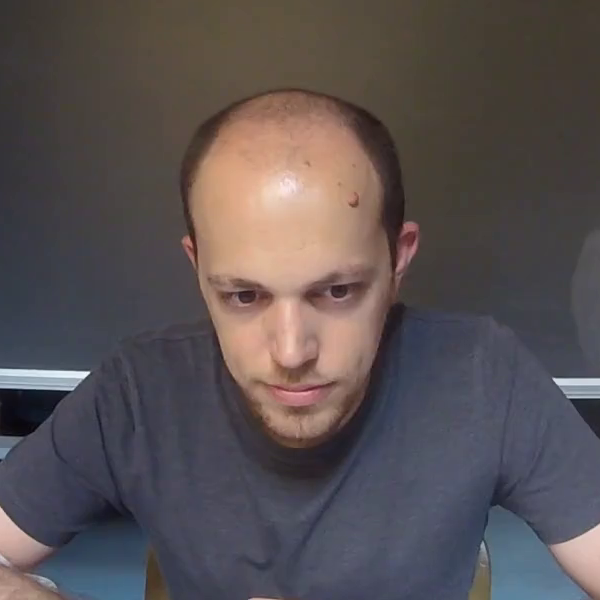}
		\includegraphics[width=\myw]{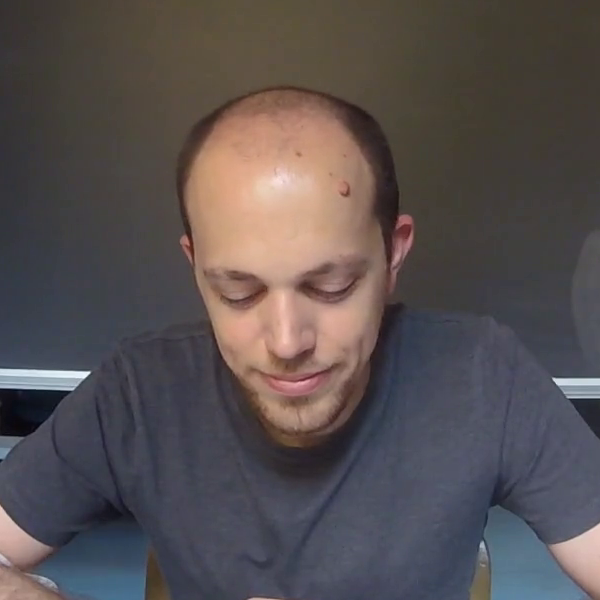}
		\includegraphics[width=\myw]{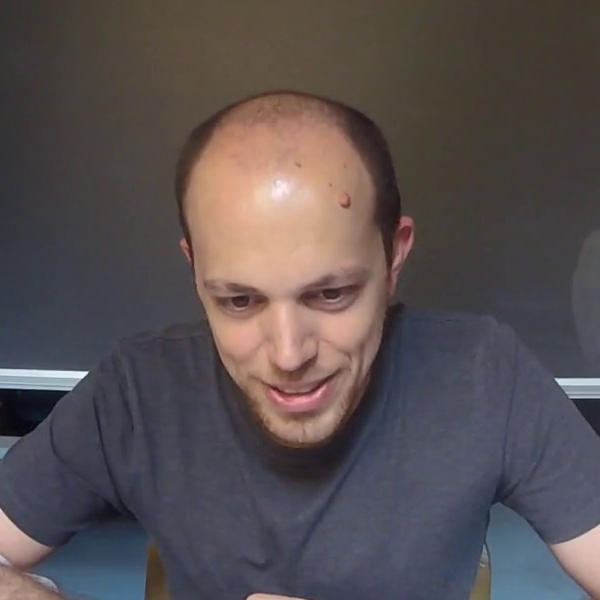} 	
		\includegraphics[width=\myw]{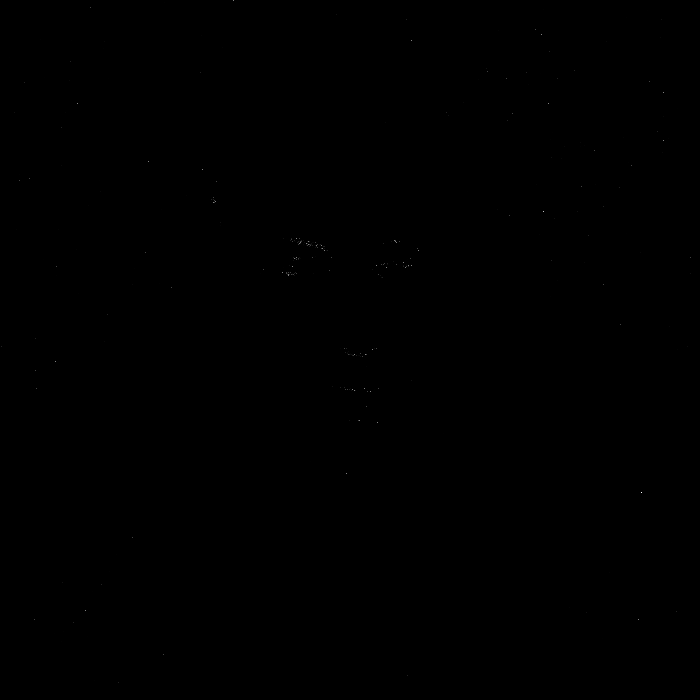}
		\includegraphics[width=\myw]{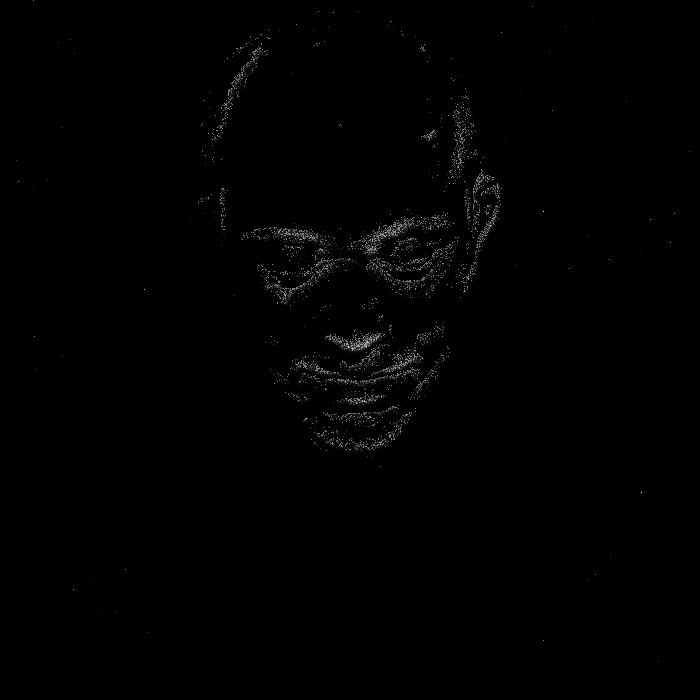}
		\includegraphics[width=\myw]{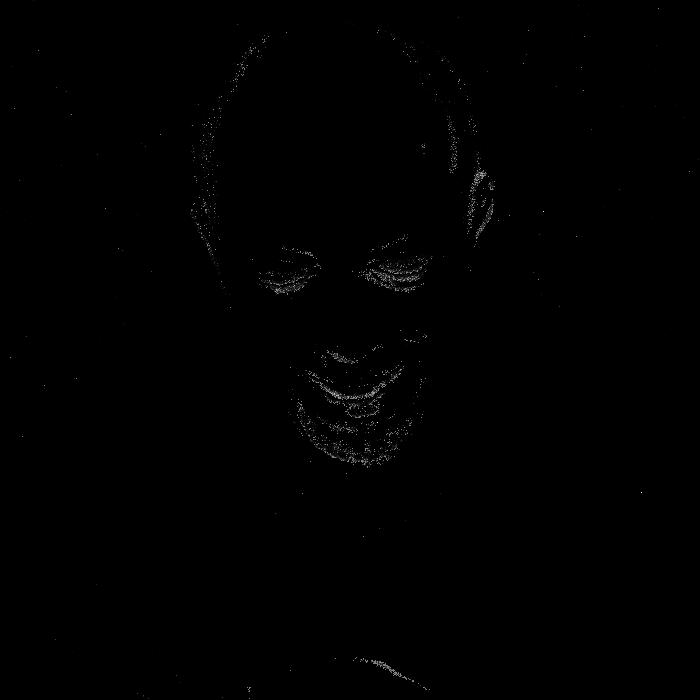} \\

  		\includegraphics[width=\myw]{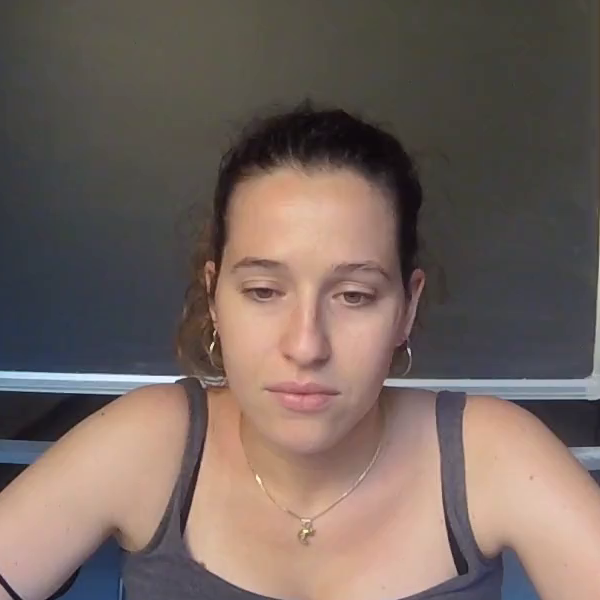}
		\includegraphics[width=\myw]{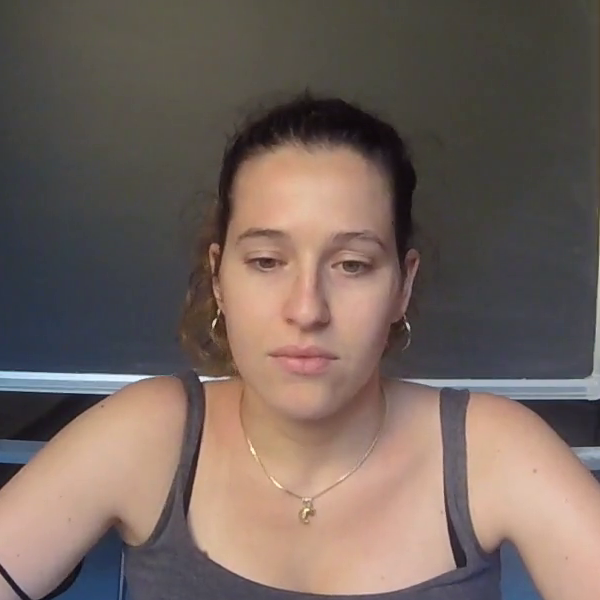}
		\includegraphics[width=\myw]{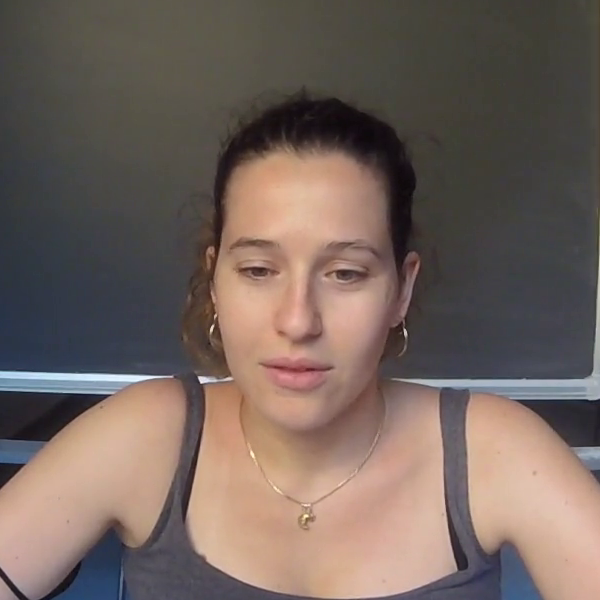} 
		\includegraphics[width=\myw]{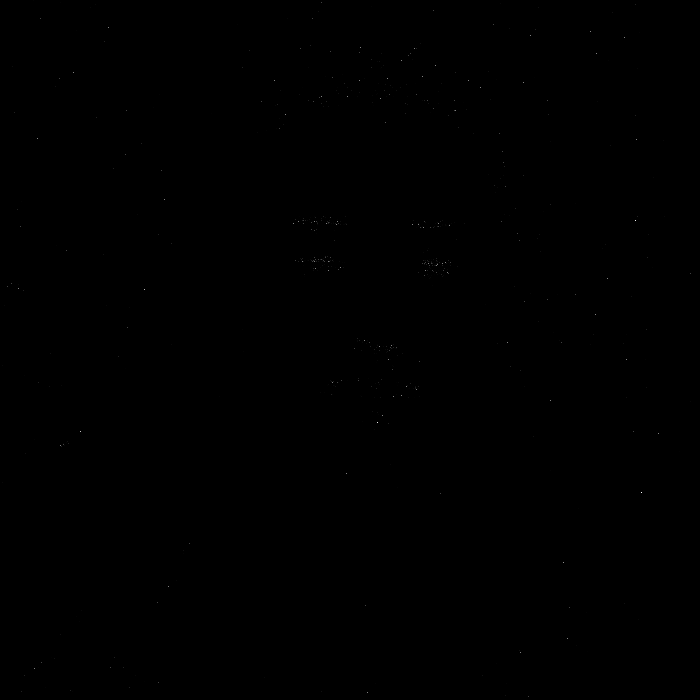}
		\includegraphics[width=\myw]{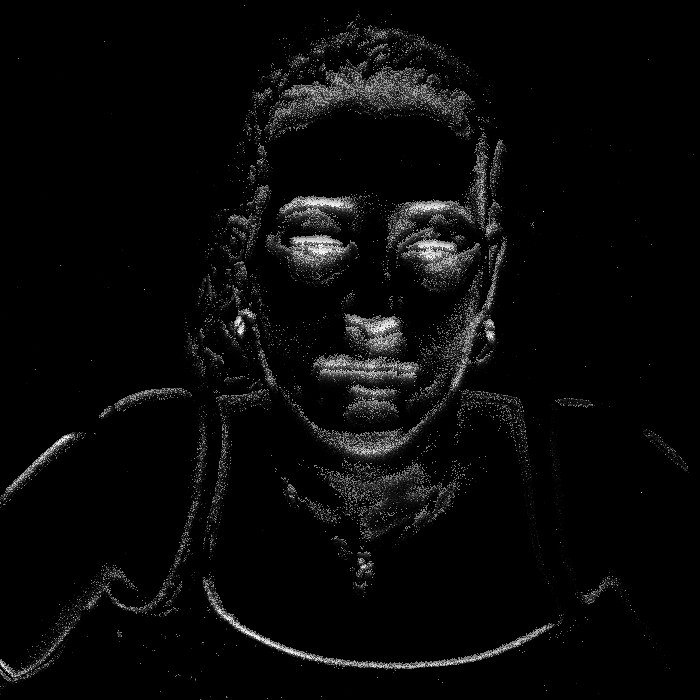}
		\includegraphics[width=\myw]{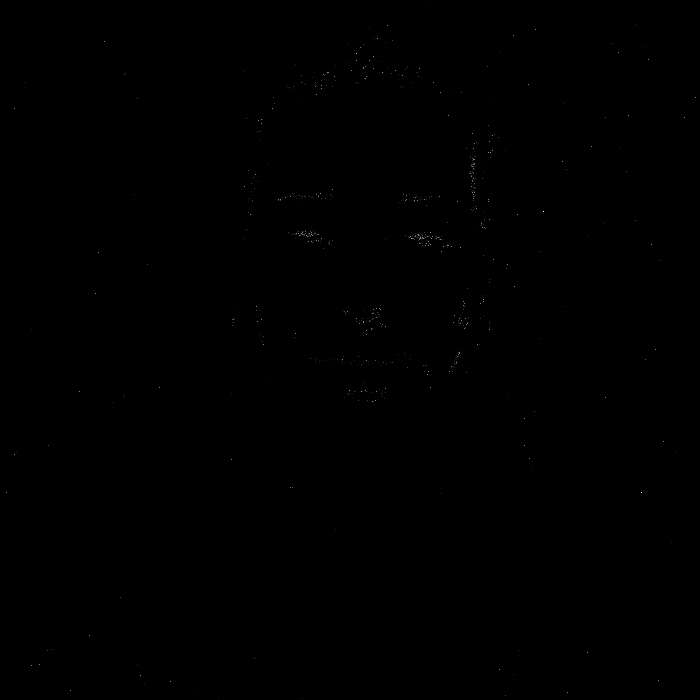} \\

		\includegraphics[width=\myw]{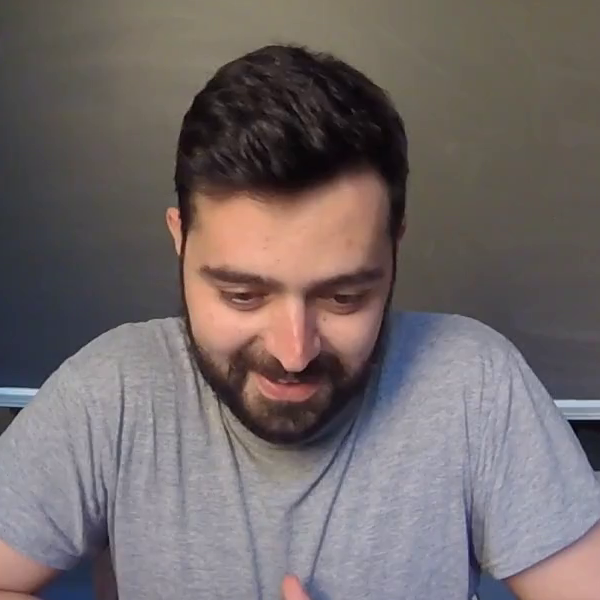}
		\includegraphics[width=\myw]{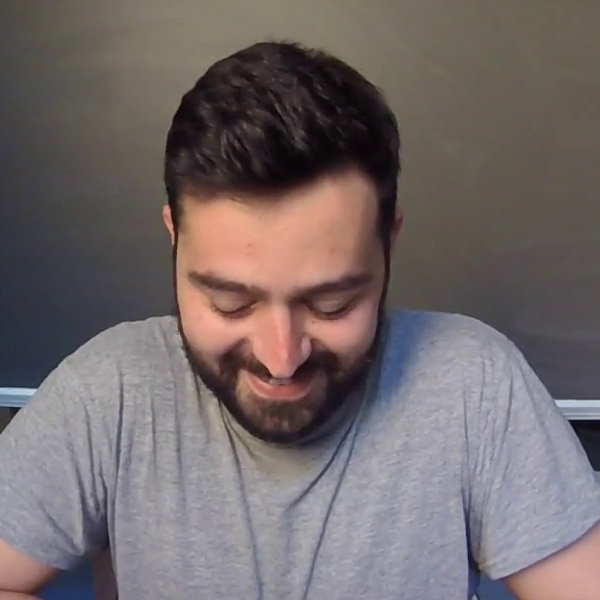}
		\includegraphics[width=\myw]{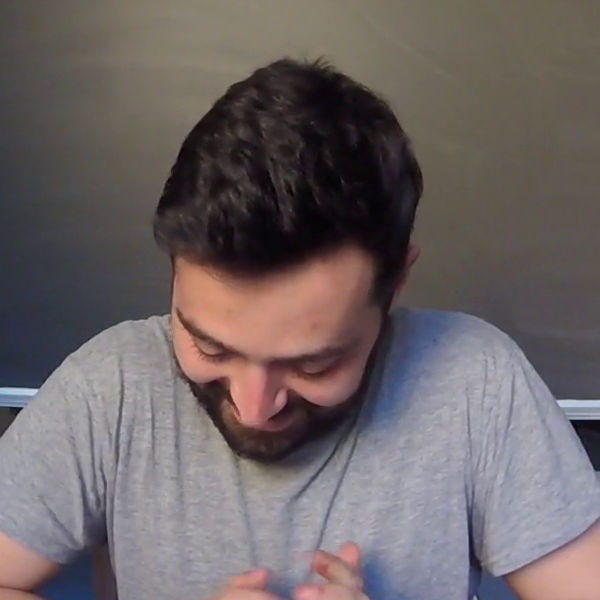}
		\includegraphics[width=\myw]{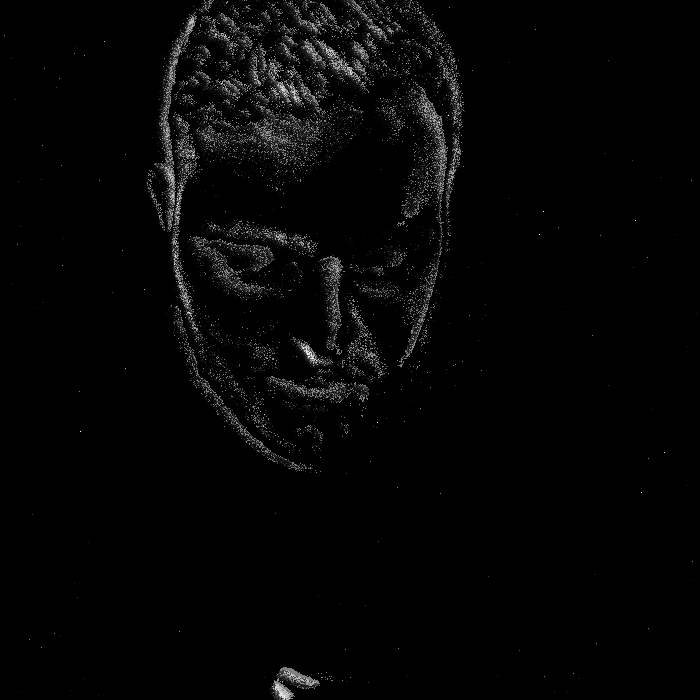}
		\includegraphics[width=\myw]{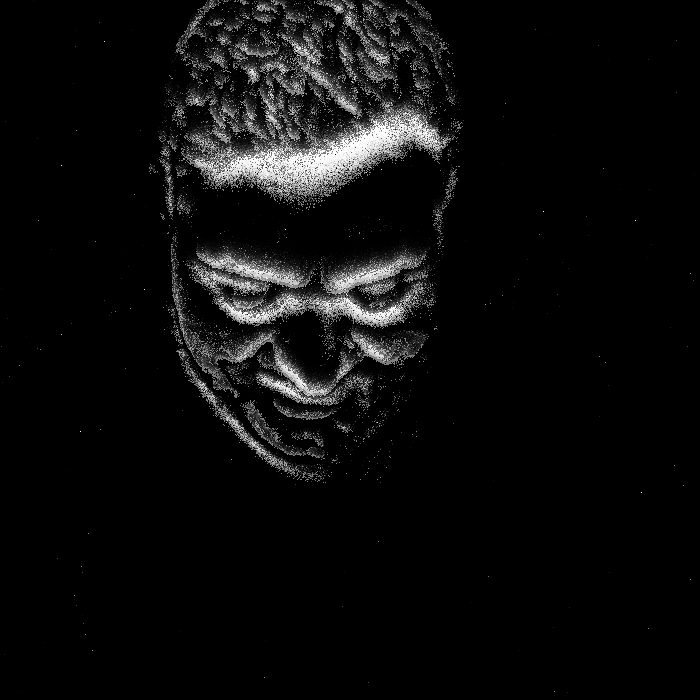}
		\includegraphics[width=\myw]{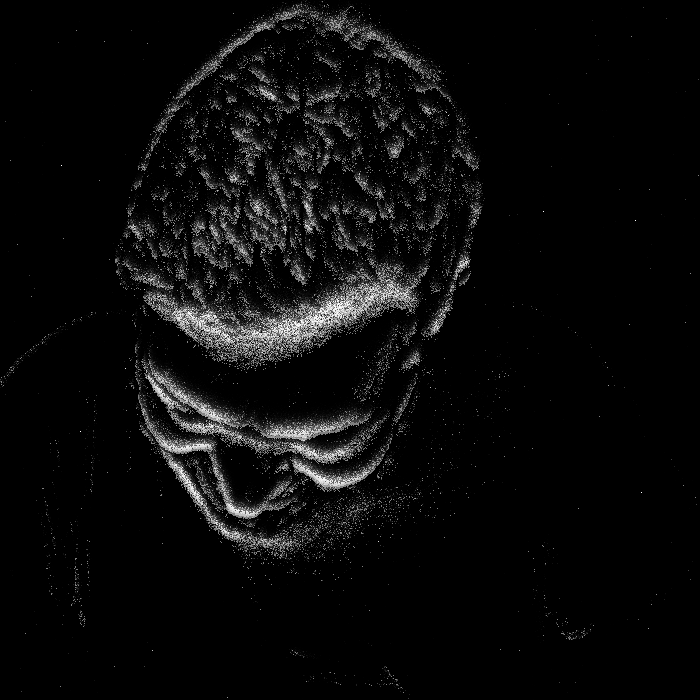} \\

            \includegraphics[width=\myw]{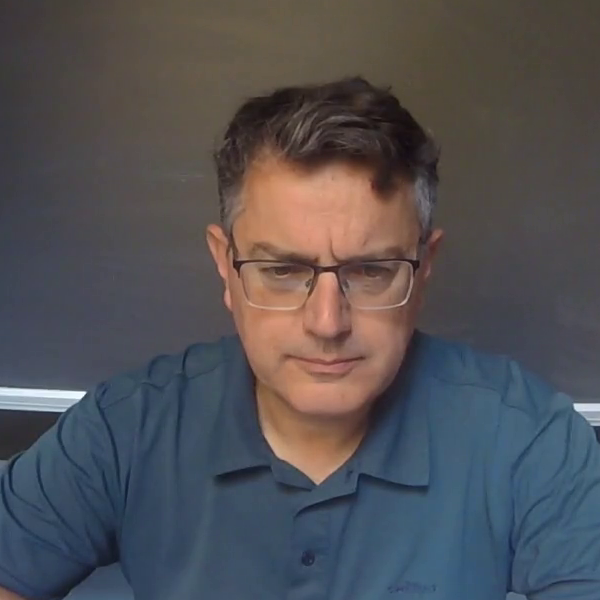}
		\includegraphics[width=\myw]{img/event_microexpr/rgb/11/0/frame_16.png}
		\includegraphics[width=\myw]{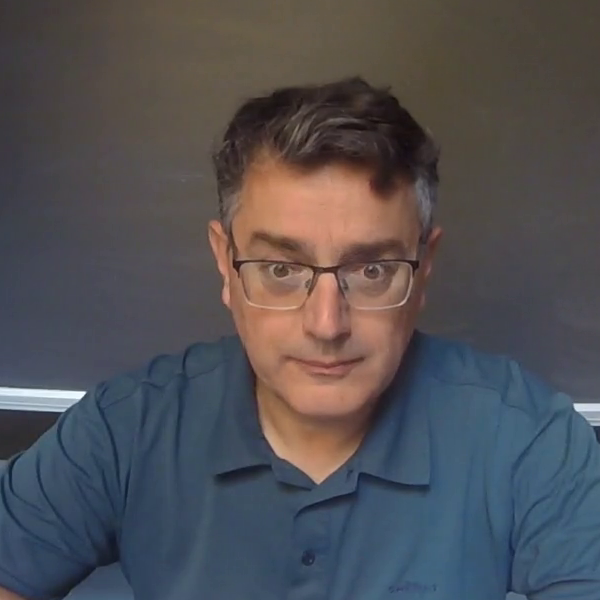} 
		\includegraphics[width=\myw]{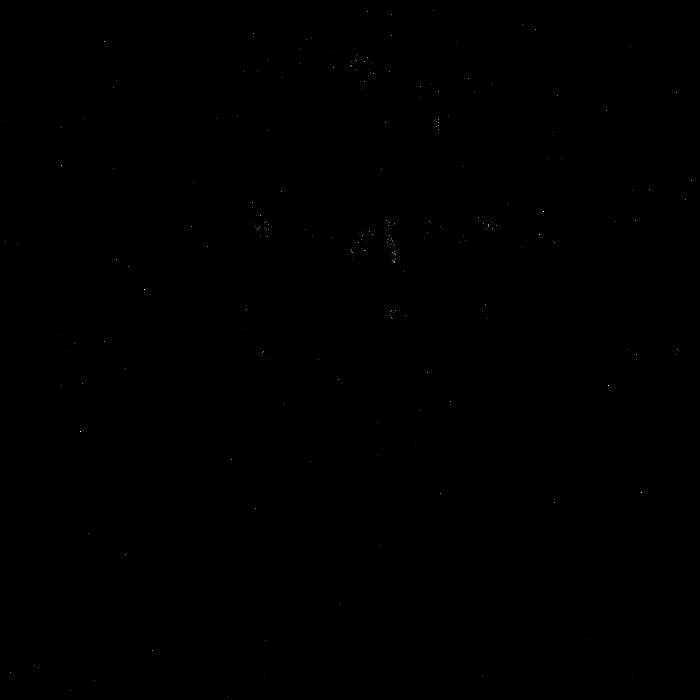}
		\includegraphics[width=\myw]{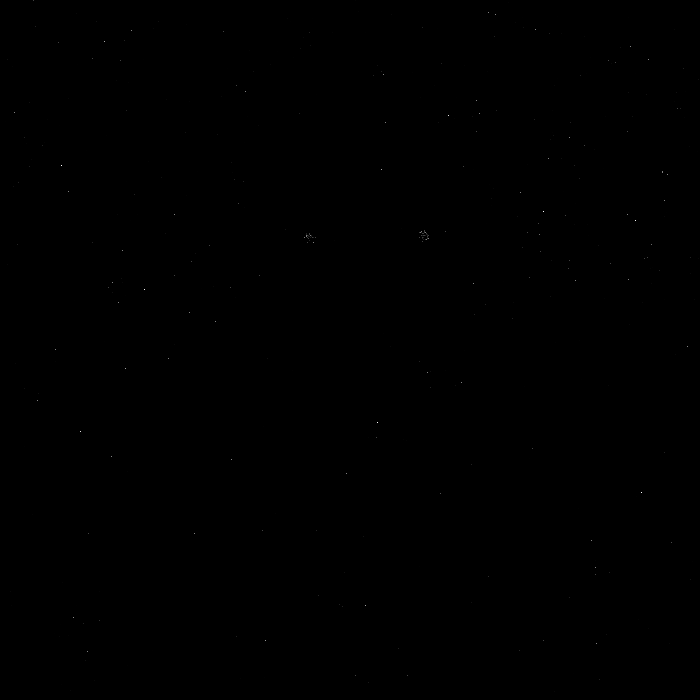}
		\includegraphics[width=\myw]{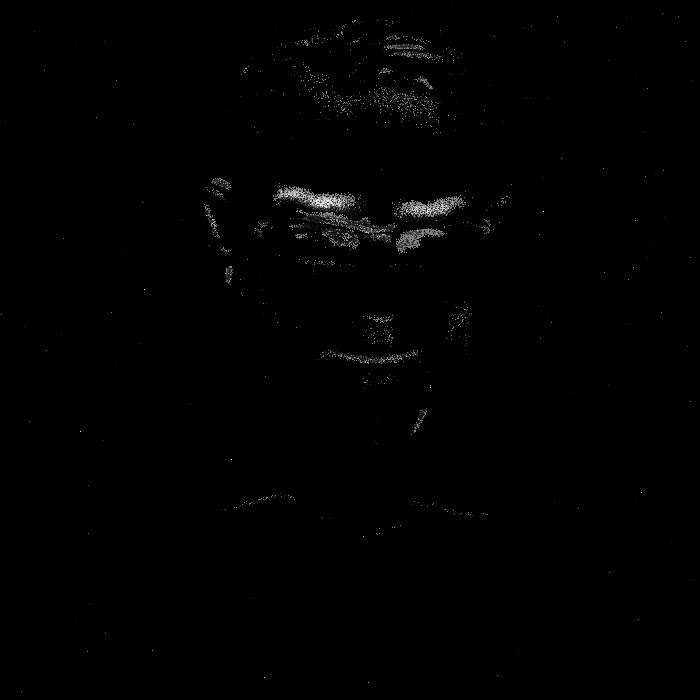} \\
	\end{center}
	\caption{Four samples from the NEFER dataset. First row: happiness; Second row: fear; third row: disgust; fourth row: surprise. Subtle movements are almost invisible with RGB but are emphasized in event frames.}
	\label{fig:samples}
\end{figure*}

For the recording we used two capturing devices: a GOPRO Hero+ action camera, recording videos at 60FPS and 1920 $\times$ 1080px resolution, and a Prophesee Evaluation Kit HD, recording event videos at a resolution of 1280 $\times$ 720px. 
The cameras have been mounted on a fixed recording rig in a room lit with natural light. We specifically avoided any presence of artificial light to avoid background noise that could alter the event-based recordings. Users are also isolated from other people which could generate distractions. 


Users have been asked to sit in front of the screen at approximately 60cm from the cameras.
The RGB and event streams have been programmatically synchronized in order to capture two videos of the same duration and content.
After viewing each video, we asked the volunteers to provide a personal evaluation of the observed footage. In particular, we asked two questions: (i) select among the 7 basic emotions, plus a \textit{"None"} option, the most suitable one to describe the emotions stemmed from viewing the video; (ii) the intensity, on a 1 to 5 scale, of such emotion.
We used the collected answers to create two alternative versions of the annotations, one considering the labeling of the user and one following our a-priori video-emotion assignment. In Fig.~\ref{fig:cf_matrix} a confusion matrix is presented showing the differences between the two label versions.
The two versions mostly differ in the fact that following user labelings we have the additional neutral emotion and a slight unbalance in the sample distribution as shown in Fig.~\ref{fig:distribution_labels}.
Overall, recording sessions lasted 18 minutes on average.
Fig.~\ref{fig:samples} shows a few samples from the dataset.

\newcommand{\wdl}{0.19\linewidth}
\begin{figure*}[t]
\begin{center}

\includegraphics[width=\wdl]{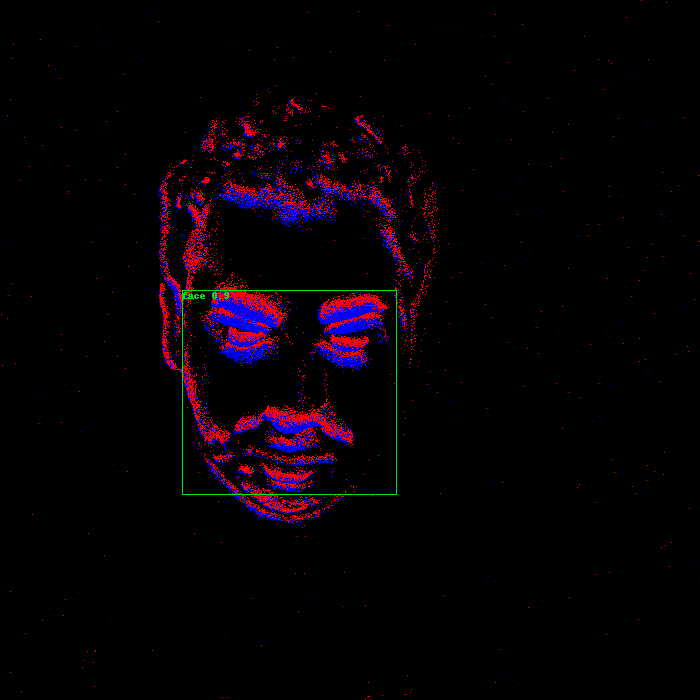}
\includegraphics[width=\wdl]{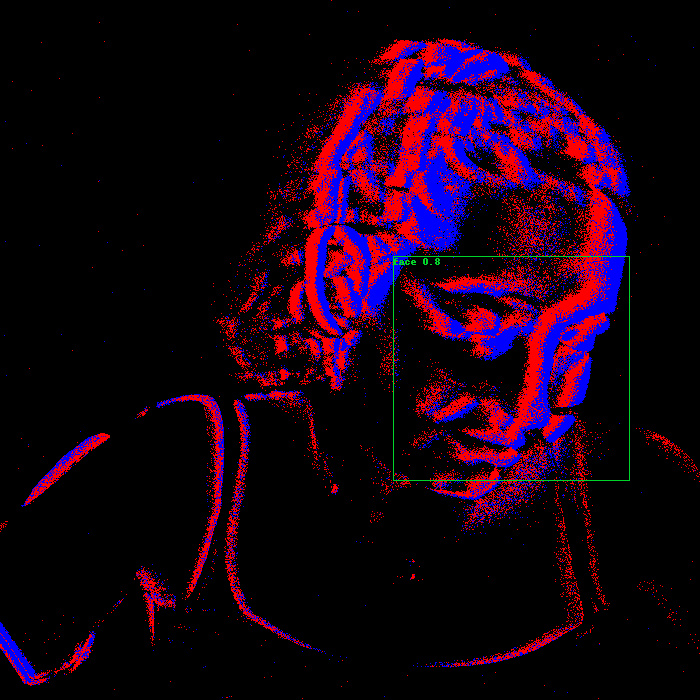}
\includegraphics[width=\wdl]{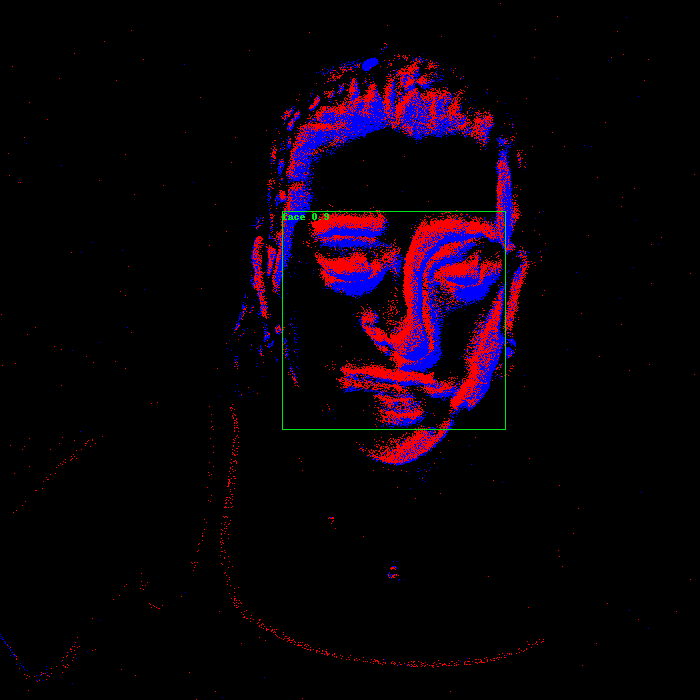}
\includegraphics[width=\wdl]{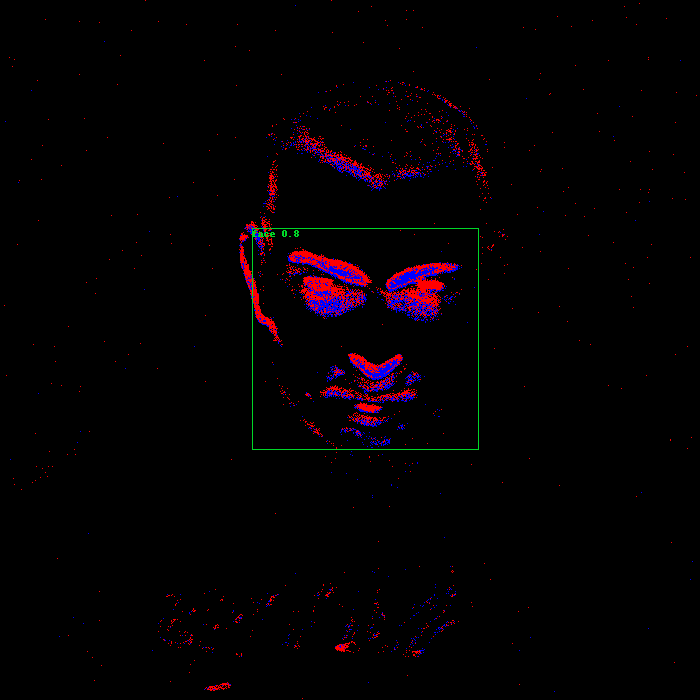}
\includegraphics[width=\wdl]{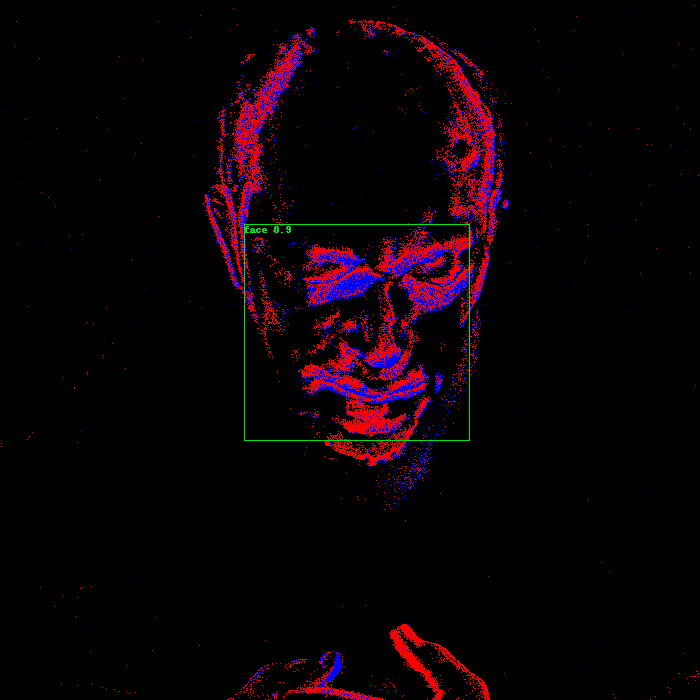} \\

\includegraphics[width=\wdl]{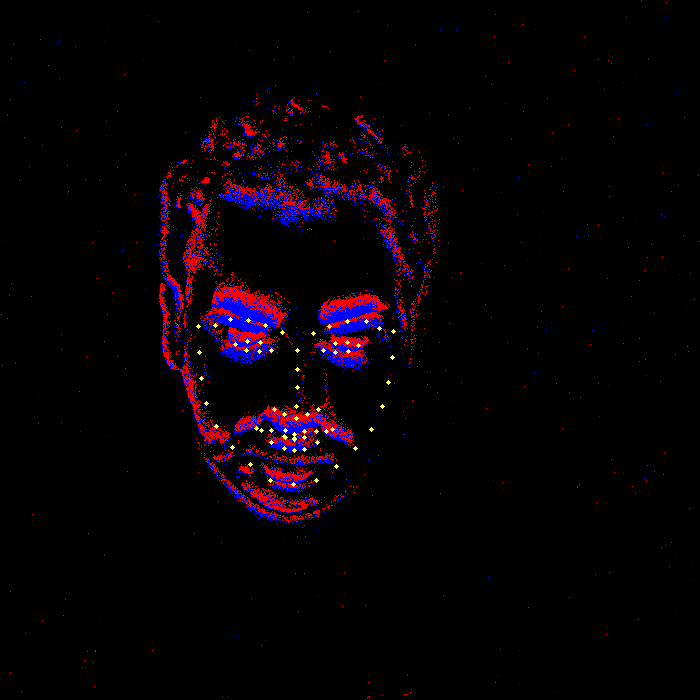} 
\includegraphics[width=\wdl]{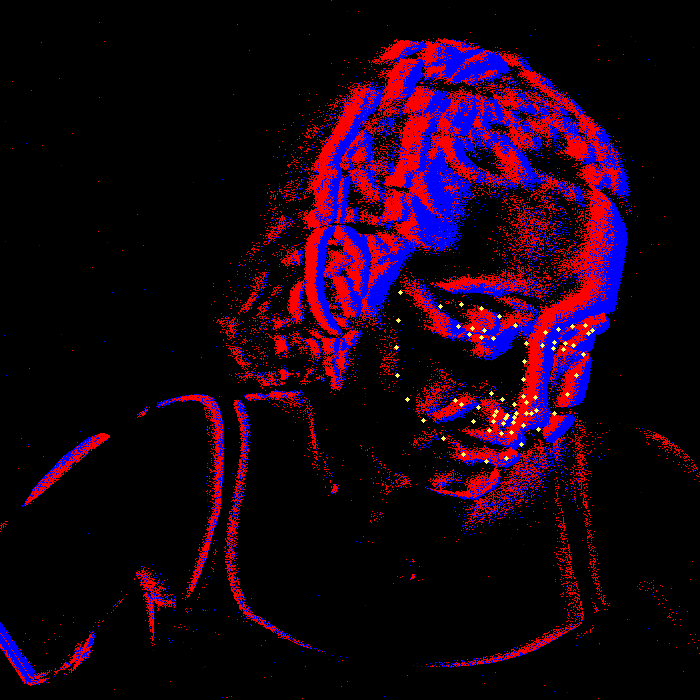} 
\includegraphics[width=\wdl]{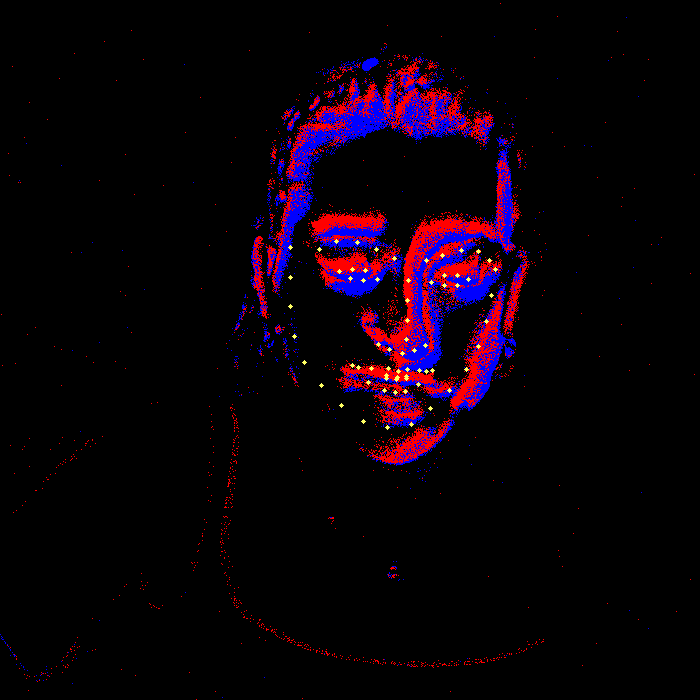} 
\includegraphics[width=\wdl]{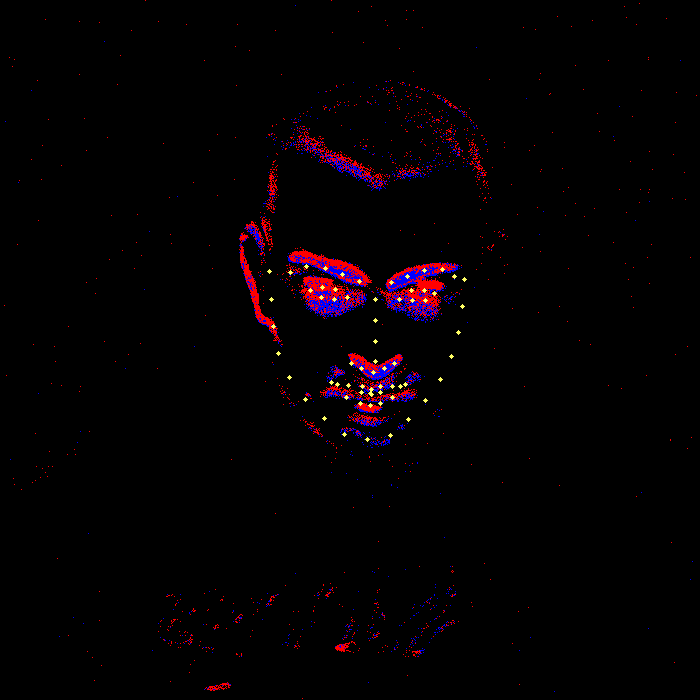} 
\includegraphics[width=\wdl]{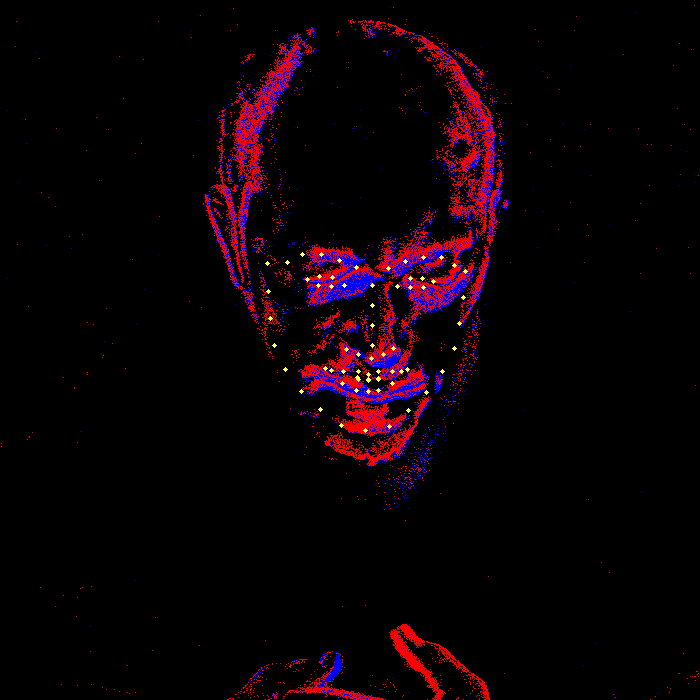} \\ 

\includegraphics[width=\wdl]{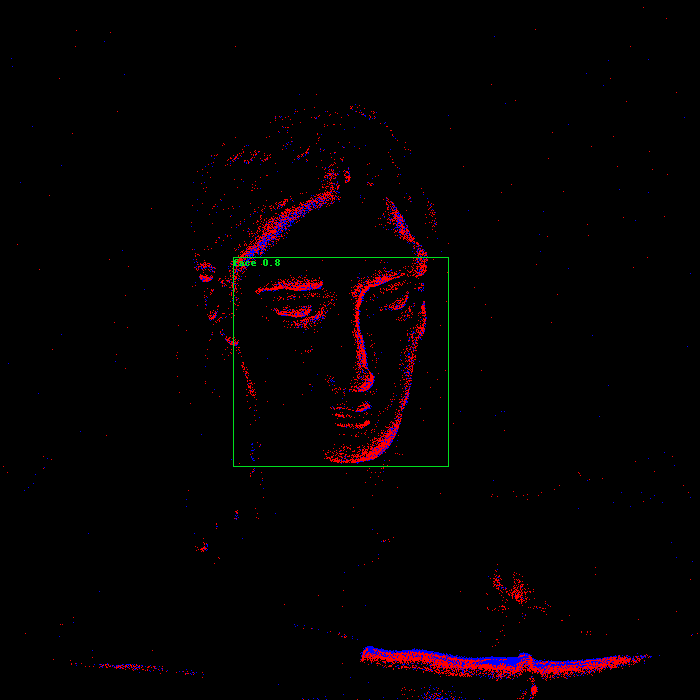}
\includegraphics[width=\wdl]{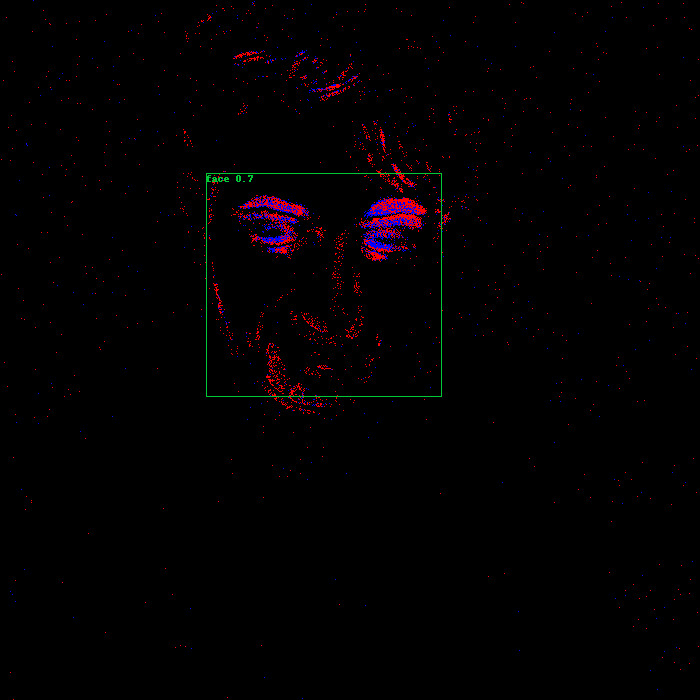}
\includegraphics[width=\wdl]{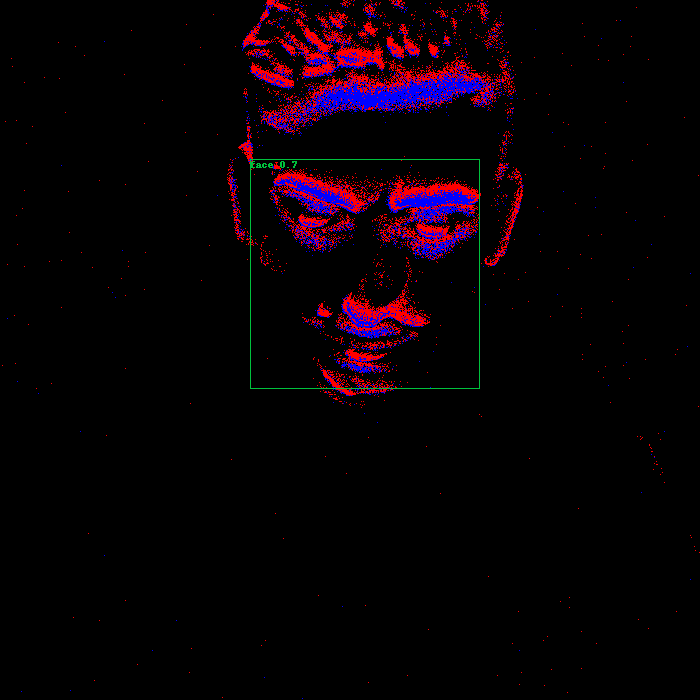}
\includegraphics[width=\wdl]{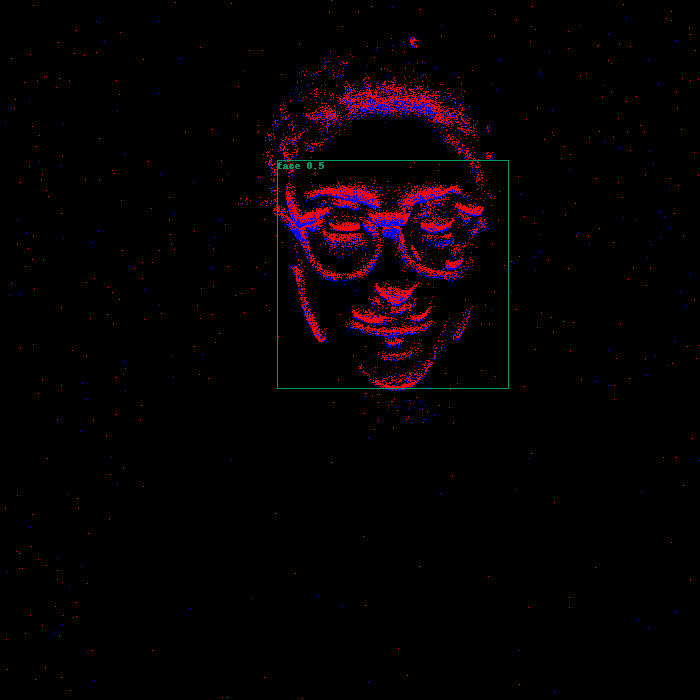}
\includegraphics[width=\wdl]{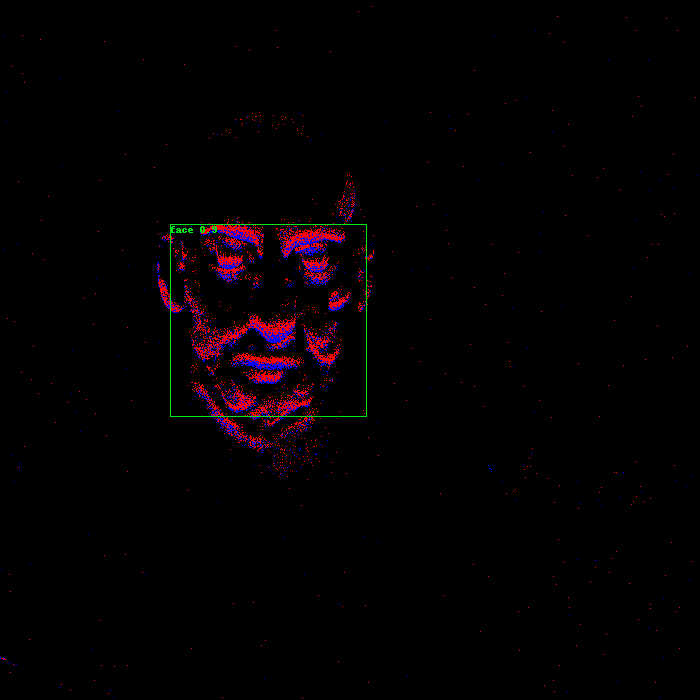}

\includegraphics[width=\wdl]{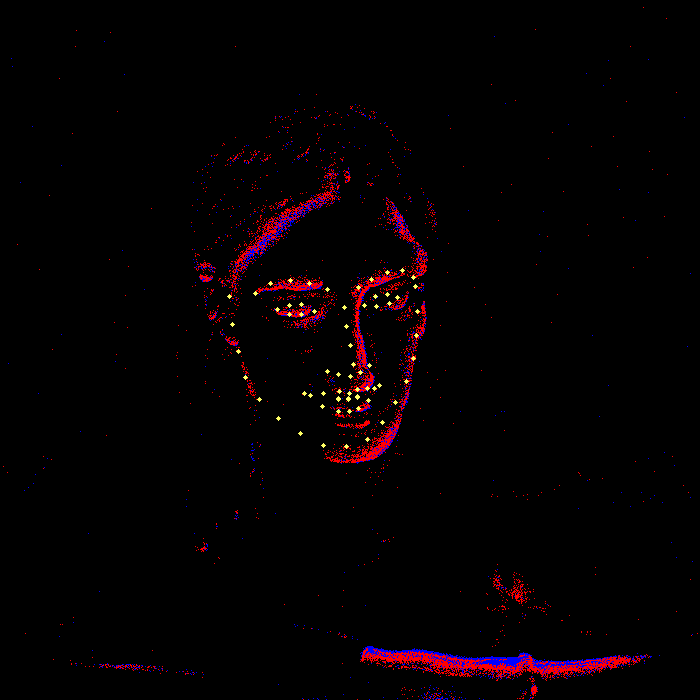} 
\includegraphics[width=\wdl]{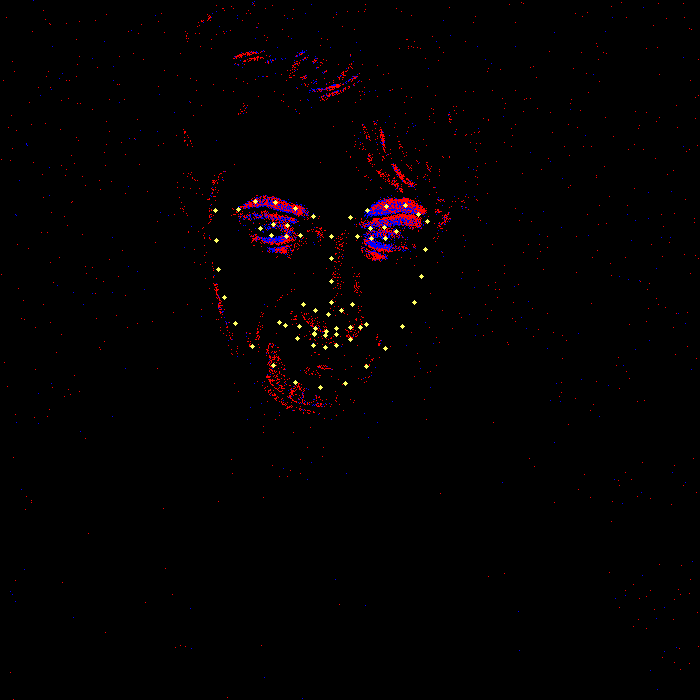} 
\includegraphics[width=\wdl]{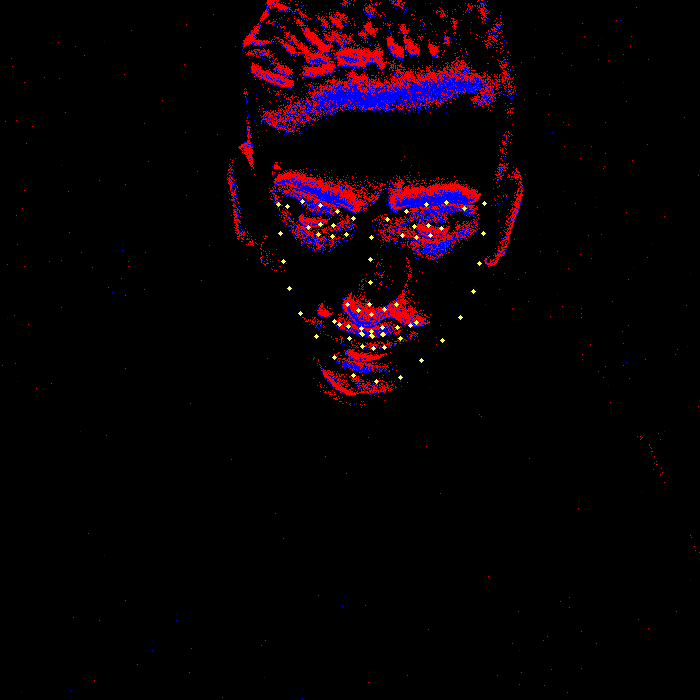} 
\includegraphics[width=\wdl]{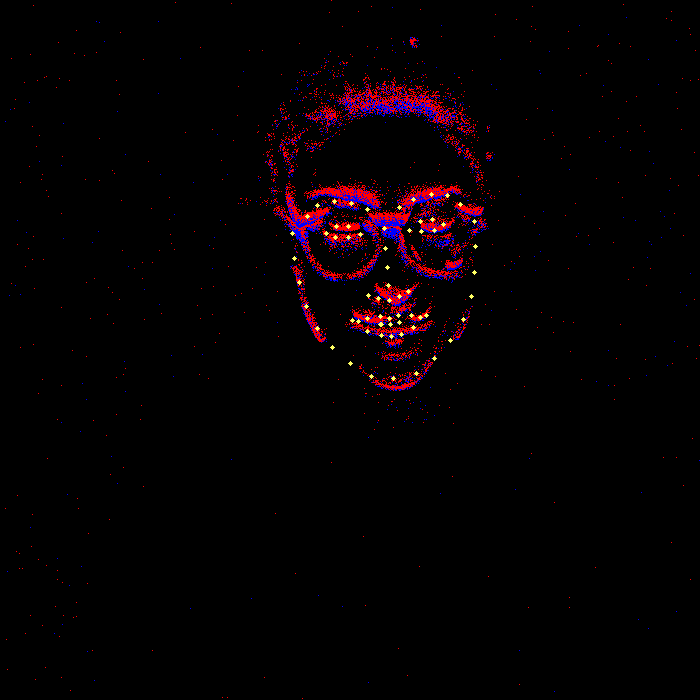} 
\includegraphics[width=\wdl]{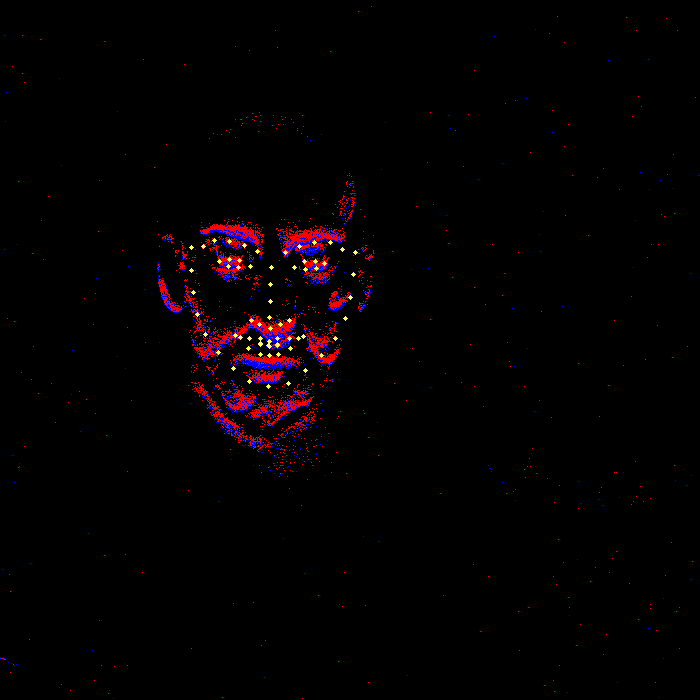}

\end{center}
  \caption{Examples of detected faces and estimated landmarks on real event videos of NEFER. Better viewed in color on a PC screen. Bounding boxes are shown in green, landmarks are shown in yellow.}
\label{fig:detland}
\end{figure*}

\section{Video Annotation Through Simulated Events}

The wide range of off-the-shelf functionalities for RGB-based computation is not available for event-based data. This includes modules that nowadays are common building blocks in computer vision pipelines such as face detectors and landmark estimators. In addition, it is necessary to preprocess the raw data of the neuromorphic sensor in order to use it with frame-based computational tools. Bridging this gap is not trivial, since due to the asynchronous nature of the domain, the usual annotation process for many different tasks becomes cumbersome and expensive.
Even generating relatively simple annotations such as facial bounding boxes, which are reliably obtainable with RGB data, would require lots of manual annotation.

To provide additional annotations for event-based data we exploit RGB data and an event camera simulator, ESIM~\cite{Rebecq18corl}. Through the use of the ESIM simulator we convert the RGB videos into physically accurate simulated event streams.
We then run a face detector and facial landmark estimator on the RGB frames, which is easily done with tools such as FaceAlignment \cite{bulat2017far}.
We train a face detector (Yolov2)\cite{redmon2017yolo9000} and a landmark estimator \cite{bulat2017far} on simulated data and test it on real event streams.
This approach provides satisfactory results on most frames, decimating the annotation time. The final annotations are manually refined and validated using CVAT \cite{vondrick2013efficiently}.

\subsection{ESIM}
ESIM~\cite{Rebecq18corl} is an event-based camera simulator that can generate a synthetic event-based stream from its RGB video counterpart in a physically realistic way.
The images are rendered by the simulator at a high frame rate, interpolating pixel brightness along the camera trajectory using an adaptive sampling technique, which is adapting the frame rate based on a prediction of the previous signals. We feed to the simulator all the RGB frames to generate a synthetic event-based version of each stream. In this way, we are able to associate the bounding boxes provided by face alignment on RGB frames with event data.
The simulator-generated outputs are encoded using an exponential time surface \cite{HOTSurface}.
Note the synthetic event-based videos obtained from the RGB data are used only as a mean for training models to quickly collect annotations. These are not pixel-wise aligned with the real event streams and we do not treat them as part of the final dataset, which only comprises real event data.

\subsection{Face Detection}\label{sub:yolo}
Using the synthetic data from the simulator, we generated an annotated dataset in the event spectrum to train a face detector.
First, we generated face annotation for RGB frames using FaceAlignment \cite{bulat2017far}, an open-source tool for face analysis\footnote{\url{https://github.com/1adrianb/face-alignment}}.
We then bound the face labels with the corresponding synthetic event frames obtained with ESIM. This allowed us to train a YOLOv2~\cite{redmon2017yolo9000} on the synthetic version of NEFER.
We found the detector to have good generalization capabilities from synthetic to real event data, which yielded high-quality annotations at a slight cost of manual validation using CVAT  \cite{vondrick2013efficiently}.

\subsection{Landmark Detection}
The facial landmark detection is performed by an Xception~\cite{chollet2017xception} architecture trained on the synthetic data from ESIM to regress the position of 68 landmarks of the face. Similarly to face detection, we obtained the ground truth labels from the RGB videos by using FaceAlignment~\cite{bulat2017far}.
The Xception architecture is composed of three stages, all of them employing depthwise separable convolutions along skip connections, resulting in a faster convergence training~\cite{chollet2017xception}. The final linear layer outputs the 136 normalized numbers representing the coordinates of the standard 68 facial landmarks.
The model is optimized using Adam with a learning rate of $8 \times 10^{-4}$ for 10 epochs over 30K frame samples with the use of standard augmentation techniques (random changes in brightness, contrast, rotation, translation, and crop).

\begin{figure*}[t]
	\begin{center}
		\includegraphics[width=0.8\linewidth]{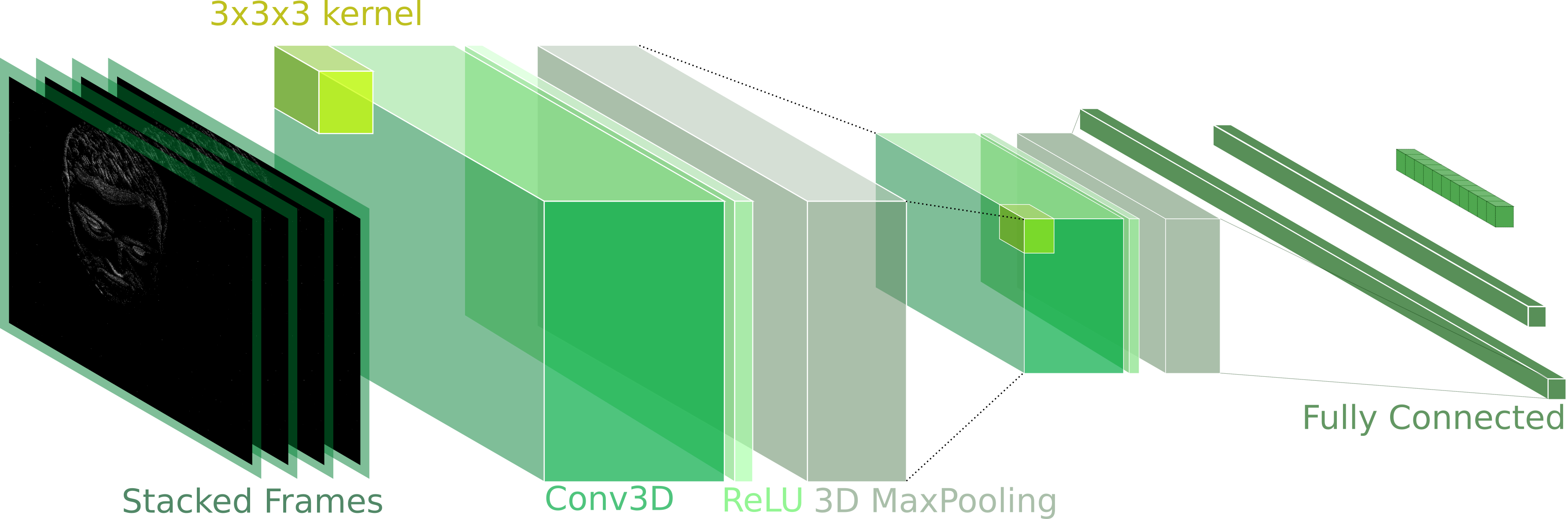}
    \end{center}
	\caption{Illustration of our C3D model. The stacked frames form the input to the first of 5 Conv3D + ReLU + 3DMaxPooling blocks. Finally the 3D feature maps are flattened and fed to the final two fully connected layers}
	\label{fig:tmp_arch}
\end{figure*}

\section{Baseline Method}
We provide a simple baseline for the dataset. This baseline architecture is based on a 3D convolutional network C3D~\cite{tran2015learning}. It has been chosen as it has been a long-standing, simple, standard approach for video-based action and activity recognition tasks\cite{Liu_Liu_Gan_Tan_Ma_2018,10.1145/2993148.2997632,montes2016temporal,tran2015learning}.
The C3D model is implemented using 5 3D convolutional blocks, all with kernel size 3 and padding 1, followed by a 3D max-pooling of size 2 and stride 2.
This chain of sequential blocks reduces the input stacked sequence of images down to a 72 channels feature map, which is then flattened and fed to two fully connected layers of size 512 and 64 before a final classification layer. ReLU activations are present between all layers. The model architecture is depicted in Fig.~\ref{fig:tmp_arch}.

We train the same model separately with RGB-frame-based data and with event data obtained by converting events into frame-wise representations using Temporal Binary Representation (TBR) \cite{innocenti2021temporal} (see Sec. \ref{tbr}). We detect the face using our pre-trained detector (see Sec. \ref{sub:yolo}), and resize the bounding box to a 200 $\times$ 200px patch before feeding it as input to the model. 

\begin{figure}[t]
\begin{center}
  \includegraphics[width=.9\linewidth]{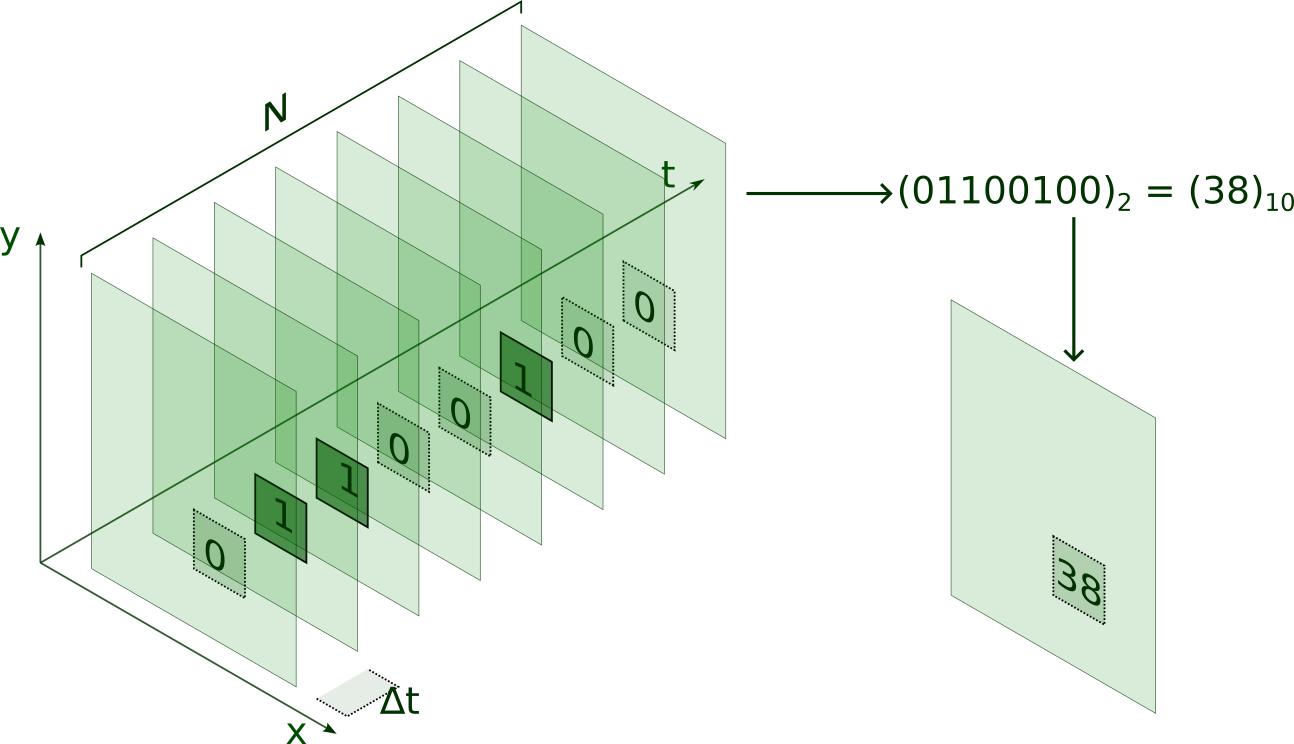}
\end{center}
  \caption{Visual diagram illustrating the TBR encoding aggregating multiple events in a frame.}
\label{fig:tbr}
\end{figure}

\subsection{Temporal Binary Representation} \label{tbr}
Temporal Binary Representation~\cite{innocenti2021temporal} (TBR) is an aggregation strategy to map the asynchronous events into a stream of synchronous frames that can be then processed by a standard computer vision pipeline. Given a fixed $\Delta t$ we can build the binary representation $b^i$ of a pixel at $(x,y)$ by checking for an event in such a time interval, $b^{i}_{x,y} = \mathbbm{1}(x,y)$.

We can then collect N consecutive representations and stack them together as $B \in \mathbb{R}^{H \times W \times N}$ forming for each pixel a binary string $[b^{0}_{x,y},b^{1}_{x,y}, ... , b^{N}_{x,y}]$, as shown in Fig.\ref{fig:tbr}.
This approach manages to create a frame processable by traditional Computer Vision algorithms with a minimal memory footprint and by retaining temporal information within the value of each pixel.

For our experiments, we used this representation setting $\Delta t = 15 $ milliseconds and $N = 8$.

\begin{figure}[t]
\begin{center}
  \includegraphics[width=.95\linewidth]{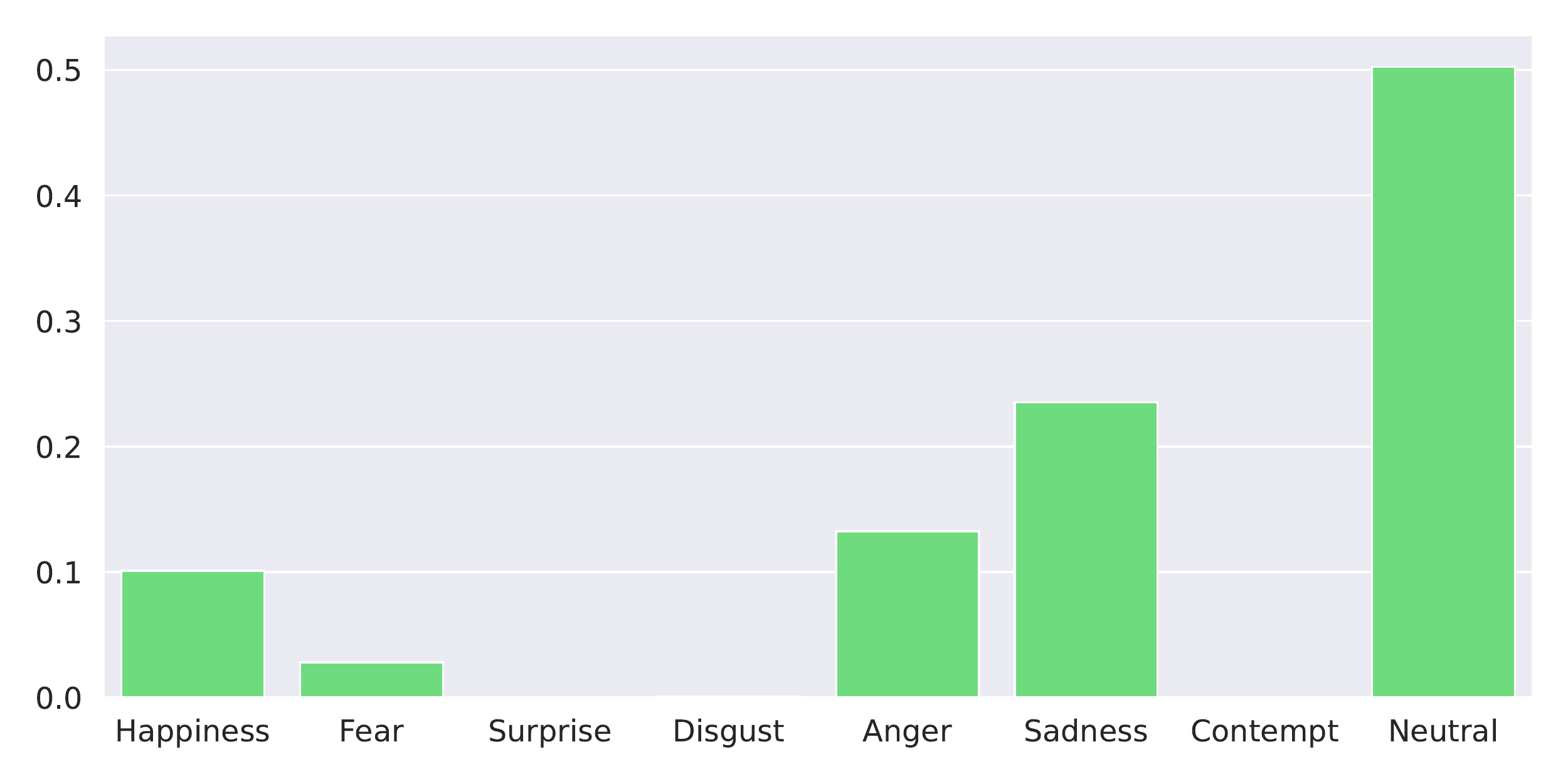}
\end{center}
  \caption{Distribution of predicted labels on frames of the NEFER validation set using Deep Face \cite{serengil2021lightface}. Almost 50\% of the frames are predicted as neutral, whereas \textit{Surprise}, \textit{Disgust} and \textit{Contempt} are predicted for less than 0.1\% of the frames.}
\label{fig:deepface_labels}
\end{figure}

\section{Experimental Results}
We implemented our C3D model using PyTorch and trained it using the Adam optimizer initialized at the default learning rate value of $1 \times 10^{-4}$ which is then reduced following the scheduling technique presented in \cite{smith2019super} with the annealing strategy.
As loss, we adopt the Binary Cross-Entropy Loss, regularized with weight decay.

We compare the performances of our model by training it separately first on the RGB videos and then on the event streams, using both the self-reported user annotations and the a-priori expected one as labels for the target emotion. We define a validation split by selecting 20\% of the users at random (thus keeping each user either in the training set or in the validation set to avoid unwanted biases), for a total of 126 videos.

We found that the RGB model results in poor accuracy, obtaining an average of 14.37\% using the user labels and 14.60\% using the expected ones.
The event-based model instead showed much better performances, reaching an accuracy of 22.95\% with the user labels and of 30.95\% using the expected ones. 
We report these experimental results in Tab. \ref{tab:accu}.
This confirms that neuromorphic cameras are well suited for analyzing faces and that event footage carries valuable information for identifying subtle micro-expressions that are not easily detectable with RGB data.

Interestingly, we observed that our baseline model, just as the human a-priori assumptions, tends to confuse classes that share similar expressions, such as \textit{fear} with \textit{surprise} or \textit{anger} with \textit{contempt} even when trained on the self-reported emotions.

Finally, we perform a control experiment by running a frame-based pre-trained state of the art emotion recognition framework on the RGB data. As a model we adopt Deepface~\cite{serengil2021lightface}, a recent facial attribute analysis framework. The model uses the same categories as we do, following Ekman's emotion classification, with the only exception of the \textit{Contempt} category, which is missing in Deepface.
As shown in Fig.\ref{fig:deepface_labels}, we note its tendency towards classifying most of the frames with the neutral class \textit{None}. This underlines the difficulty of the task in the setting that we propose: most frames do not carry a very polarized expression and most emotion cues happen very quickly, in a way that it is difficult to grasp them with RGB cameras. We argue that to fully comprehend the underlying emotions of humans from a vision-based point of view, event cameras will play an important role in the near future due to their ability to capture fine-grained micro-expressions and micro-movements of the face.

\begin{table}[t]
\resizebox{0.48\textwidth}{!}{%
\begin{tabular}{c|cc|cc} 
\hline
 Data & A-Priori Labels  &  \%  & Reported Labels  &  \% \\ 
 \hline \hline
   RGB & 14.60 &   - & 14.37 &   - \\ \hline
   TBR Event & 22.95 & +57.2\% & \textbf{30.95} &  +115.4\% \\ \hline

 \end{tabular}}
 \caption{Absolute accuracy and relative performances of our baseline model over the different data domains and using both labelling versions of NEFER. \label{tab:accu}}
\end{table}

\section{Conclusions and Future Work}
In this paper, we presented a first release of NEFER, a dataset for expression recognition based on event camera data. This dataset is composed of paired visual spectrum images and event camera streams. For every sequence of frames, both the expected emotion and the self reported one by the user are given. Every frame has multiple annotations, namely the user face bounding box and the respective facial landmarks that we collected by leveraging models trained on synthetic data obtained using a simulator.
Finally, we presented and discussed a 3D convolutional baseline, trained on both version of our dataset, which achieved improved results on event camera data with respect to the RGB frame based data.

We consider this a starting point for a future larger collection of data in the event camera domain for similar high-time resolution tasks. The large interest given by the computer vision community towards understanding facial expressions and emotions proves the importance of the task, yet the neuromorphic community and the traditional RGB vision one still have several gaps to be bridged. We believe that pursuing this line of research will bring attention to an emerging field, bringing together the best of both worlds and providing multiple modalities to approach problems that, based on experimental results, appear to be better addressed in the event domain rather than in the RGB domain alone.

\section*{Acknowledgements}
This work was supported by the European Commission under European Horizon 2020 Programme, grant number 951911—AI4Media.

{\small
\bibliographystyle{ieee_fullname}
\bibliography{egbib}
}

\end{document}